\documentclass[sigconf]{acmart}
\usepackage{subcaption}
\usepackage{colortbl}
\usepackage{multirow}

\graphicspath{figures/}

\usepackage[capitalize]{cleveref}
\crefname{section}{Sec.}{Secs.}
\Crefname{section}{Section}{Sections}
\Crefname{table}{Table}{Tables}
\crefname{table}{Tab.}{Tabs.}

\newcommand{\ie}{\textit{i}.\textit{e}.}

\usepackage{lipsum}

\AtBeginDocument{%
  \providecommand\BibTeX{{%
    \normalfont B\kern-0.5em{\scshape i\kern-0.25em b}\kern-0.8em\TeX}}}

\copyrightyear{2023}
\acmYear{2023}
\setcopyright{acmlicensed}
\acmConference[MM '23] {Proceedings of the 31st ACM International Conference on Multimedia}{October 29--November 3, 2023}{Ottawa, ON, Canada.}
\acmBooktitle{Proceedings of the 31st ACM International Conference on Multimedia (MM '23), October 29--November 3, 2023, Ottawa, ON, Canada}
\acmPrice{15.00}
\acmISBN{979-8-4007-0108-5/23/10}
\acmDOI{10.1145/3581783.3612018}

\begin{document}

\title{Ada3Diff: Defending against 3D Adversarial Point Clouds via Adaptive Diffusion}

\author{Kui Zhang}
\email{zk19@mail.ustc.edu.cn}
\affiliation{%
  \institution{University of Science and Technology of China}
  \country{}
}
\author{Hang Zhou}
\authornote{Corresponding authors.}
\email{zhouhang2991@gmail.com}
\affiliation{%
  \institution{Simon Fraser University}
  \country{}
}
\author{Jie Zhang}
\email{jie_zhang@ntu.edu.sg}
\affiliation{%
  \institution{Nanyang Technological University}
  \country{}
}

\author{Qidong Huang}
\email{hqd0037@mail.ustc.edu.cn}
\affiliation{%
  \institution{University of Science and Technology of China}
  \country{}
}

\author{Weiming Zhang}
\authornotemark[1]
\email{zhangwm@ustc.edu.cn}
\affiliation{%
  \institution{University of Science and Technology of China}
  \country{}
}

\author{Nenghai Yu}
\email{ynh@ustc.edu.cn}
\affiliation{%
  \institution{University of Science and Technology of China}
  \country{}
}

\renewcommand{\shortauthors}{Kui Zhang et al.}
% \affiliation{%

% \affiliation{%

% \affiliation{%

% \affiliation{%

% \affiliation{%


\begin{abstract}
 Deep 3D point cloud models are sensitive to adversarial attacks, which poses threats to safety-critical applications such as autonomous driving. Robust training and defend-by-denoising are typical strategies for defending adversarial perturbations. However, they either induce massive computational overhead or rely heavily upon specified priors, limiting generalized robustness against attacks of all kinds. To remedy it, this paper introduces a novel distortion-aware defense framework that can rebuild the pristine data distribution with a tailored intensity estimator and a diffusion model. To perform distortion-aware forward diffusion, we design a distortion estimation algorithm that is obtained by summing the distance of each point to the best-fitting plane of its local neighboring points, which is based on the observation of the local spatial properties of the adversarial point cloud. By iterative diffusion and reverse denoising, the perturbed point cloud under various distortions can be restored back to a clean distribution. This approach enables effective defense against adaptive attacks with varying noise budgets, enhancing the robustness of existing 3D deep recognition models. 
\end{abstract}






\maketitle

\section{Introduction}
Recently, an increasing amount of studies have demonstrated tremendous success in using deep models for large-scale 3D point cloud analysis, and with the easy affordability of 3D sensors, recognition models for 3D point clouds have been deployed in various applications such as autonomous driving, virtual reality, and robotics.
However, it has been demonstrated that deep models can be easily misled by adversarial point clouds. For example, with a carefully designed optimization strategy ~\cite{cw,cw-knn,geoa3,siadv,advpc,aof}, the attacker can directly shift points on 3D surfaces to generate adversarial point clouds, which can preserve the original shapes to bypass human inspection and can even transfer attack ability across different models.
These attacks pose a significant threat to safety-critical applications.
\begin{figure}[t]
\begin{center}
  \includegraphics[width=0.94\linewidth]{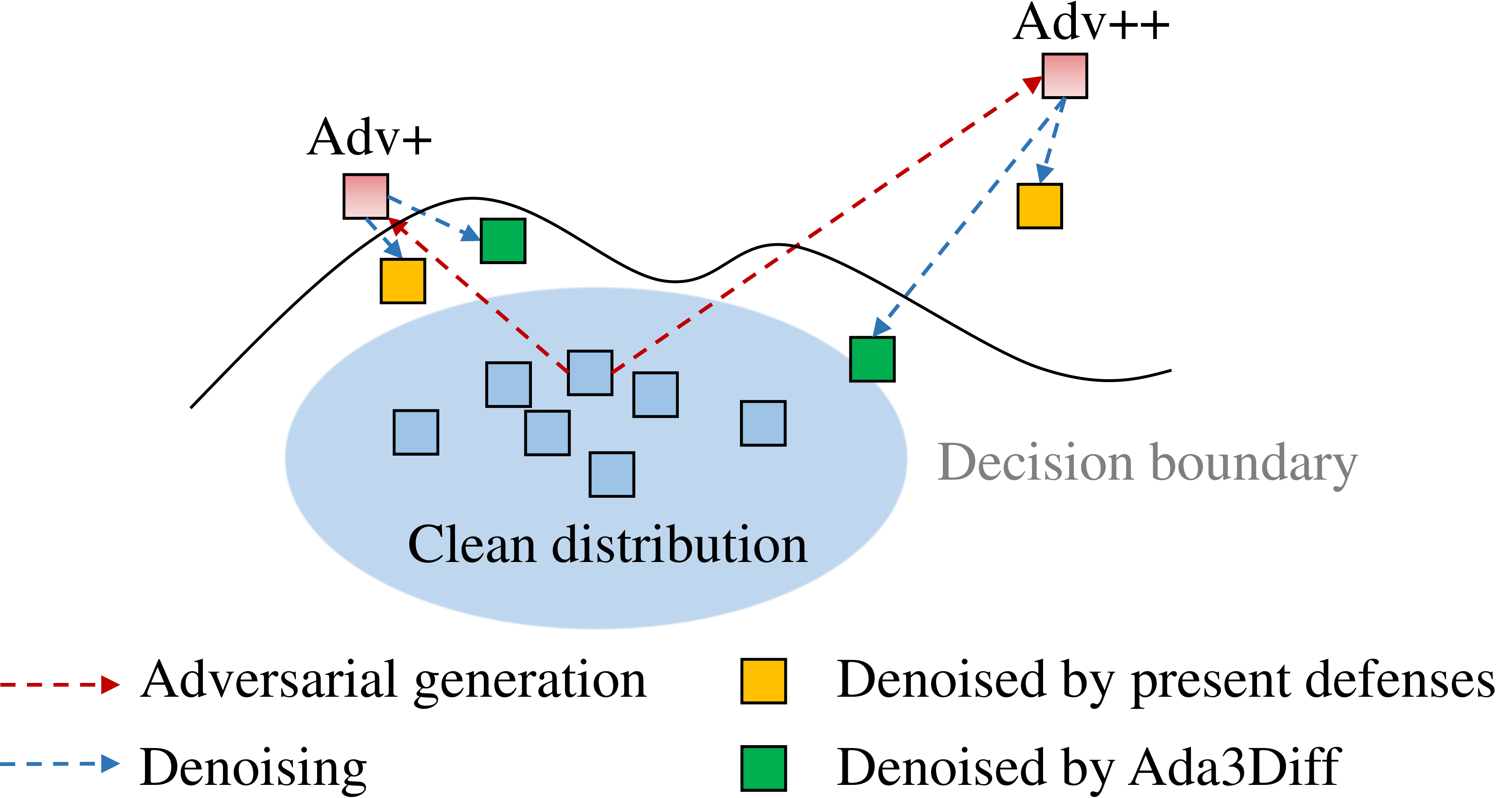}
\end{center}
  \caption{Comparison of our Ada3Diff with present defenses qualitatively.
  For varying distorted adversarial point clouds, Ada3Diff obtains denoised point clouds closer to the clean distribution than others. Adv+ and Adv++ denote the adversarial attacks with slight and large distortions, respectively. }
\label{fig:teaser}
\vspace{-1mm}
\end{figure}

To mitigate this issue, two main defense strategies have been proposed. One is dedicated to training a robust classifier to capture discriminative features for analysis ~\cite{xiaoyi,jiachen,LPC}. Commonly, one can obtain a robust classifier by fine-tuning the original classifier or training it from scratch, both of which will introduce massive computational costs and suffer from low accuracy.
The other one is to append a pre-processing module before the well-trained model to eliminate the effect of adversarial noises~\cite{dupnet,if-defense}. Despite easy plug-in-and-play and model-agnostic properties, it is inapplicable for denoising adversarial point clouds under different noise budgets as it depends heavily on specified spatial prior.
In Figure \ref{fig:teaser}, we demonstrate the difference between present defenses and our defense in high-level feature space. Current defenses are limited by their inability to effectively restore the adversarial point cloud with large distortions to clean distribution.
Taking denoised-based defense, IF-Defense~\cite{if-defense}, as an example, we find that it achieves a satisfying performance on point clouds with slight distortions, yet it fails on large distortion cases, which inherently violates its basic assumption of preserving surface continuity for robust recognition.

In this paper, we revisit the poor generalizability of preprocessing-based defenses from the distribution perspective. We argue that limited spatial priors are inadequate for effectively guiding the denoising of complex adversarial point clouds, and therefore propose the use of distribution prior to enhancing generalizability. Specifically, we present a distribution-guided defense method, Ada3Diff, to denoise adversarial point clouds, which can intrinsically move diverse adversarial distributions back to the clean distribution. We model data distributions of clean point clouds using a diffusion probabilistic model and gradually eliminate noise from adversarial point clouds. 
We design a distortion-aware forward diffusion algorithm to disrupt potential perturbations in diverse adversarial point clouds and align them as closely as possible to the transition distribution at training, followed by iteratively reverse denoising to restore the perturbed distribution to a clean distribution. 
In the process of distribution restoration, one challenge is how to select an appropriate diffusion timestep for adversarial point clouds with different distortions, as a large timestep is not conducive to reconstructing shape-invariant adversarial point clouds while a small one fails to eliminate adversarial perturbation effectively. 
Unlike DiffPure~\cite{DiffPure}, which uses fixed timesteps for all attacks, we employ distortion-aware adaptive diffusion to overcome this challenge. 
Specifically, we compute the distance between each point and its neighborhood fitting plane and estimate the distortion of the input based on the average distance of the entire point set, which is inspired by the observation that non-shape-invariant point clouds tend to be locally chaotic. With distortion estimation, we can identify the proper diffusion step for each instance in each iteration in order to perform adaptive forward diffusion and reverse denoising.

Extensive experiments demonstrate that Ada3Diff outperforms existing defense methods in terms of robustness against attacks of most kinds, especially facing greater noise. Besides, we provide a comprehensive evaluation to showcase the effectiveness of our method on different datasets, models, and attack scenarios. 

	In summary, our contributions are three-fold: 
\begin{itemize} 

\item We find the reliance on limited spatial priors hinders the generalization of defense, and propose a general defense against point cloud attacks from a distribution perspective.
\item We propose a distortion-aware defense framework including distortion estimation and adaptive diffusion, which can effectively defend against adversarial attacks with various distortions by rebuilding the pristine data distribution. 
\item Extensive experiments show that our method outperforms current defense methods by a large margin facing large distortions, and meanwhile, it achieves comparable performance on point clouds with slight distortions. 
\end{itemize} 
\section{Related Works}
\subsection{3D Point Clouds Attacks}\label{sec:attack}
In addition to guaranteeing the attack performance, existing 3D point clouds attack methods~\cite{cw,cw-knn,geoa3,lggan,Minimal,spga,siadv,advpc,aof} target satisfying other two requirements, \ie, imperceptibility and transferability. 
For imperceptibility, the well-known gradient-based method PGD~\cite{pgd} can be adjusted for a slight distortion, while
3D-Adv~\cite{cw} can append additional distance constraints to limit perceptible perturbations. 
To support the adversarial point clouds for better surface reconstruction, $k$NN-Adv~\cite{cw-knn} introduced  an extra kNN loss for its optimization objective. 
SI-Adv~\cite{siadv} constraint points are moved tangentially along the surface to generate a shape-invariant adversarial point cloud.
Differently, Point dropping~\cite{drop} deleted the most critical points based on the saliency map.  All the above methods can generate adversarial point clouds with smooth surfaces and smaller average perturbation distances. 
To improve transferability, attack methods such as AdvPC~\cite{advpc} and AOF~\cite{aof} generated adversarial point clouds that tend to have a large degree of local disorganization, manifested by a more uniform distribution of local points within the sphere. Both attack methods sacrificed imperceptibility for transferability. 
In this paper, we utilize some of the typical methods mentioned above to evaluate robustness against adversarial point clouds with various distortions.
\begin{figure*}[htbp]
\centering
\subcaptionbox{Clean}{
\begin{minipage}[t]{0.24\linewidth}
\centering
\includegraphics[width=\linewidth]{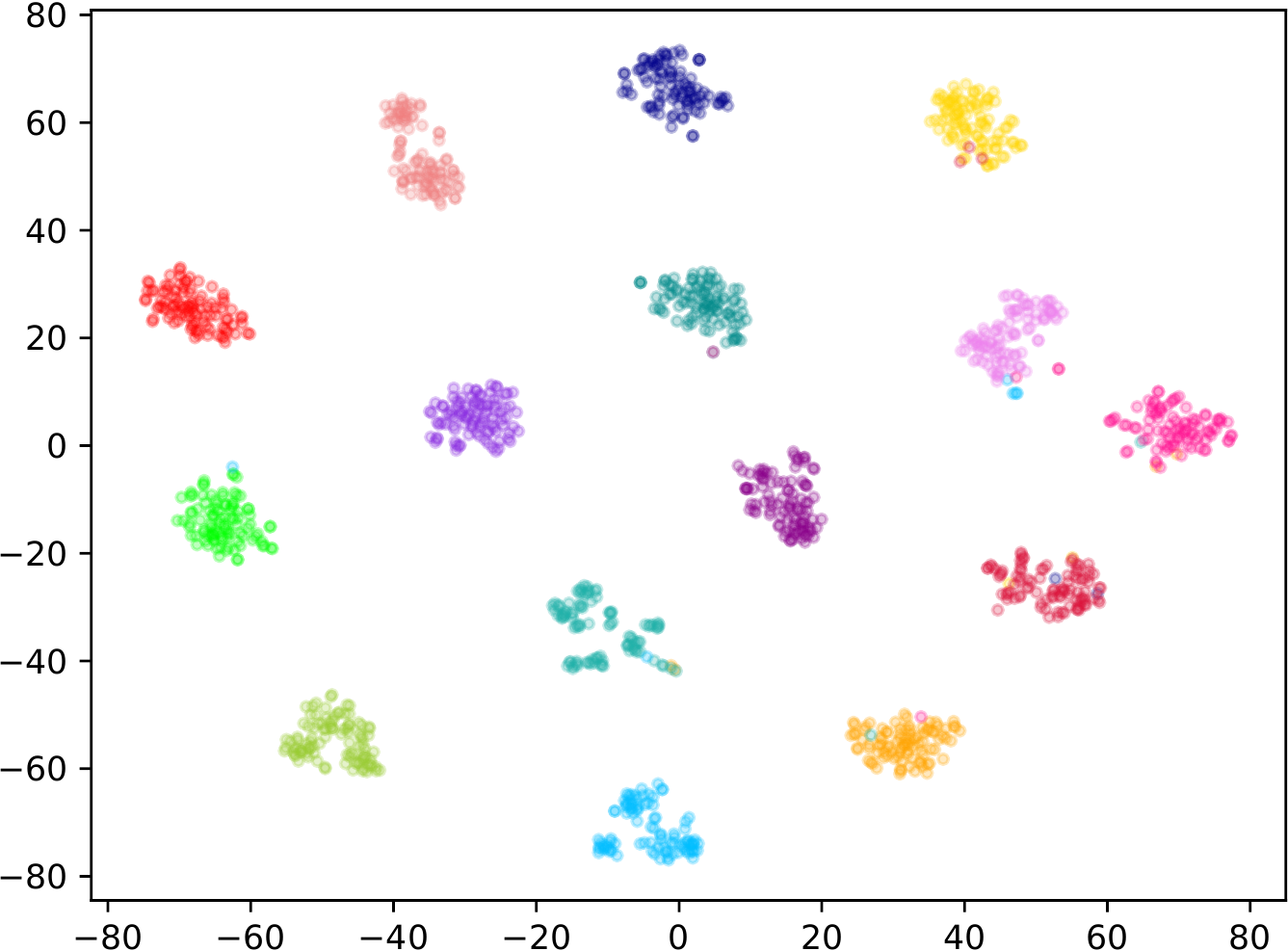}
\end{minipage}%
 }%
\subcaptionbox{AOF}{
\begin{minipage}[t]{0.24\linewidth}
\centering
\includegraphics[width=\linewidth]{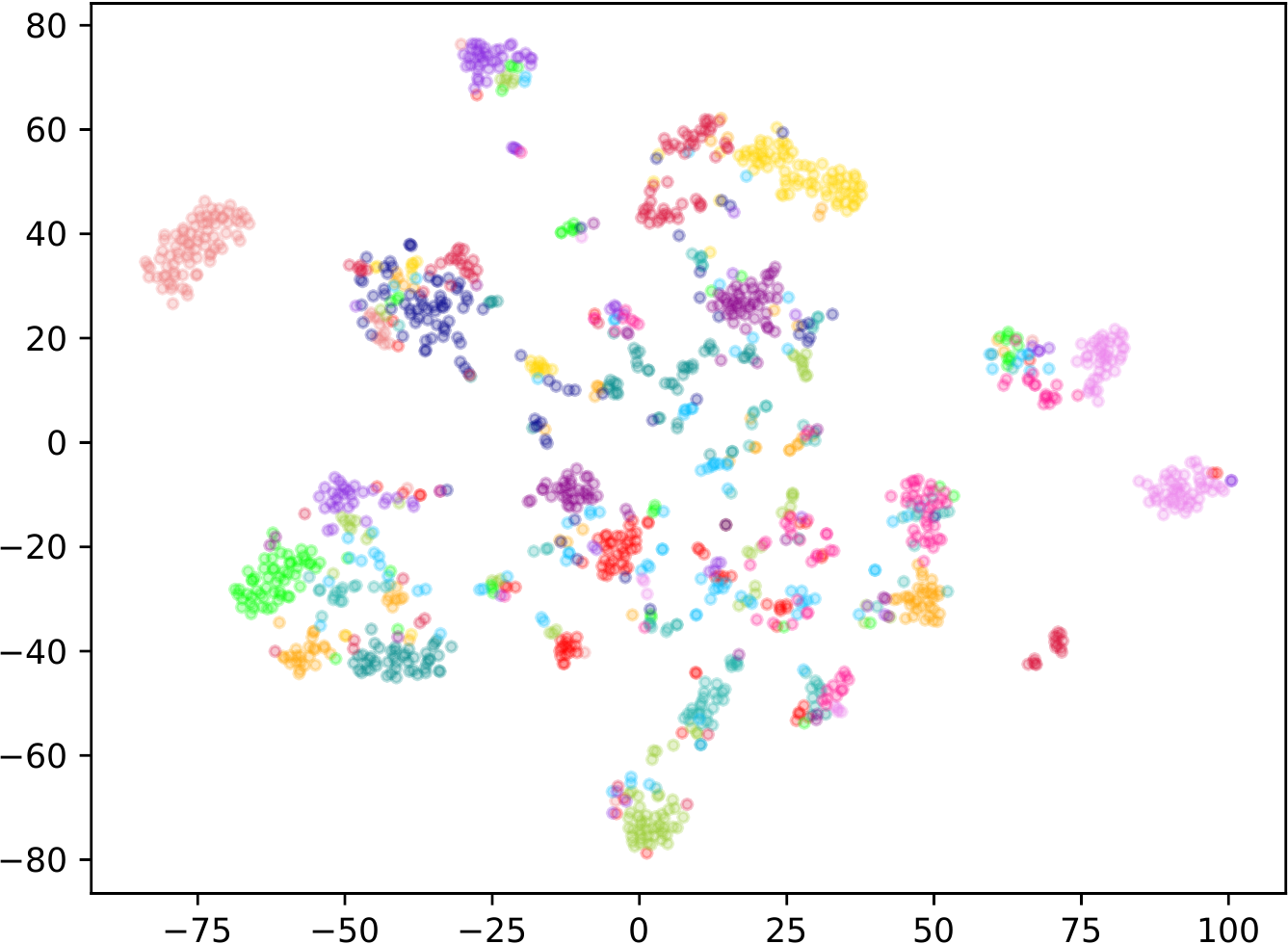}
\end{minipage}%
}%
\subcaptionbox{DUP-Net}{
\begin{minipage}[t]{0.24\linewidth}
\centering
\includegraphics[width=\linewidth]{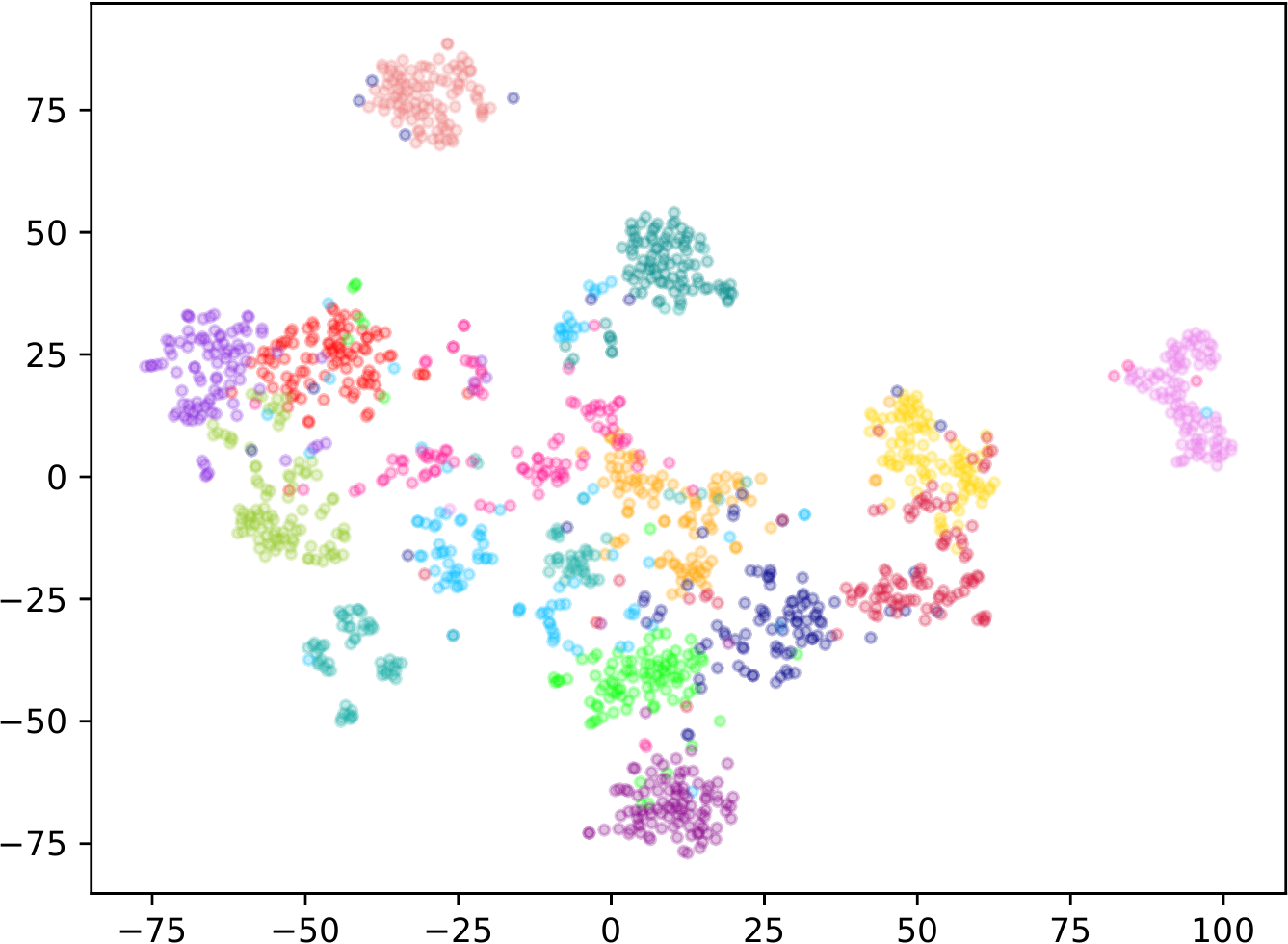}
\end{minipage}%
}%
\subcaptionbox{IF-Defense}{
\begin{minipage}[t]{0.24\linewidth}
\centering
\includegraphics[width=\linewidth]{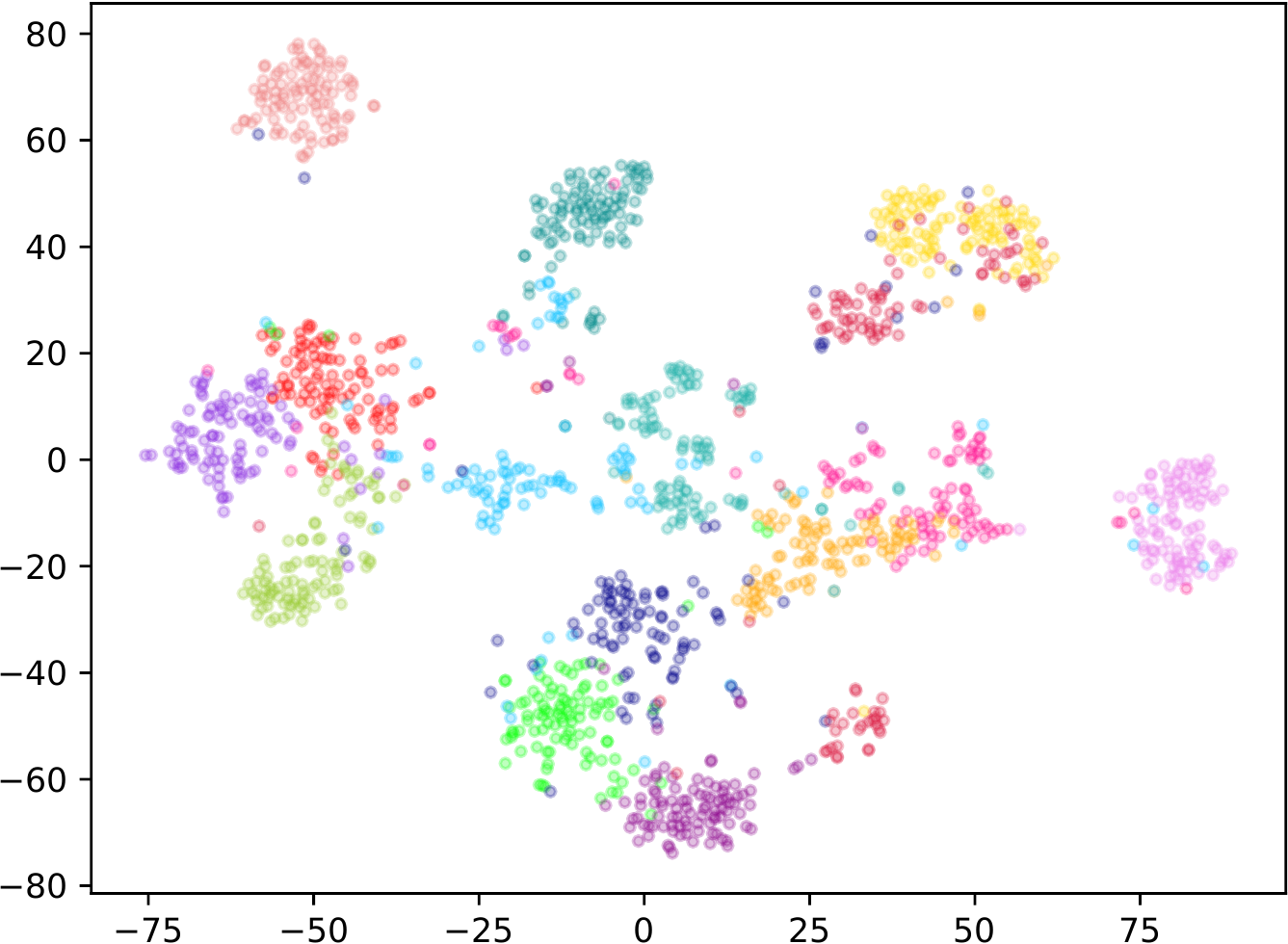}
\end{minipage}%
}%
\centering
\caption{The t-SNE visualizations of the latent space. (a) and (b) are the distributions visualized for clean and AOF adversarial point clouds. (c) and (d) are the distribution of the denoised adversarial point clouds by DUP-Net and IF-Defense respectively.}
\label{tsne:motivation}
\end{figure*}

\subsection{3D Point Cloud Defenses}
3D Point Cloud Defenses are mainly achieved by building robust classifiers or purifying adversarial point clouds. \cite{RN856,jiachen,xiaoyi,LPC} construct classification models that are robust against adversarial point clouds by designing robust training methods or designing robust recognition modules and networks. 
However, all these methods will induce high computational costs and are model-specific.  Another route is to denoise the adversarial point clouds  before feeding them to the model, which  is model-agnostic and can be combined with various robust training methods.  
Simple random sampling (SRS)~\cite{srs} is a commonly-used solution to purify adversarial point clouds. Another common method, called statistical outlier removal (SOR)~\cite{sor}, assumed the adversarial points as outliers.
Based on the SOR, DUP-Net~\cite{dupnet} further supplemented the information with an additional upsampling network. However, these methods cannot handle adversarial points that deviate too far from the clean distribution.
Afterward,
IF-Defense~\cite{if-defense} utilized implicit functions and distribution-aware loss to constrain the SOR-processed points to be distributed as evenly as possible over the surface, achieving good performance against a variety of attacks, yet it performs poorly suffering larger distortions, due to the constraints of an even surface cannot guarantee that the point clouds purified back to the natural distribution. 
A concurrent work called PointDP~\cite{pointdp} also  utilizes a diffusion probabilistic model for the defense of 3D adversarial point clouds. Broadly, PointDP employs a \textit{conditional} diffusion probabilistic model and empirically chooses a \textit{fixed} diffusion timestep to incrementally denoise the adversarial point cloud.
By comparison, Ada3Diff uses an unconditional diffusion probabilistic model as the pre-processing module and conducts multi-round distortion-adaptive diffusion denoising. 

\subsection{Diffusion Probabilistic Models}
Inspired by nonequilibrium thermodynamics, the diffusion probabilistic model~\cite{RN1249} is proposed to model the process of generating data as a gradual denoising process based on Markov chains. Recent works~\cite{diffpoint,pvd,lion} have demonstrated the success of using diffusion models to model the generation of point clouds, which pointed out that the reverse diffusion process can gradually eliminate the noise of different variances to generate point clouds similar to the clean data distribution. 
Moreover, some works~\cite{DiffPure,guided-diff} utilized diffusion models to remove adversarial noise on 2D images and achieved superior performance than previous methods, including adversarial training, demonstrating the potential of diffusion models in adversarial defense. However, they are both perturbation-independent.
In this paper, we pioneer exploring the powerful diffusion probabilistic models to defend against 3D point cloud attacks and further introduce distortion-aware diffusion for adaptive defense.

\section{Method}
\subsection{Problem Definition}
For an input point cloud $\mathbf{x} \in \mathbb{R}^{N \times 3}$ containing $N$ points, the attacker $\mathcal{A}$  aims to mislead the classifier $f$ by modifying $\mathbf{x}$ under a certain perturbation budget $\xi$:
\begin{equation} \label{eq:threat_model}
\begin{aligned}
f(\mathcal{A}(\mathbf{x})) \neq f(\mathbf{x}), s.t. \; \mathcal{B}\left(\mathcal{A}(\mathbf{x}), \mathbf{x}\right) \leq \xi,
\end{aligned}
\end{equation}
where $\mathcal{B}$ is the distance measurement for adversarial shifts or the number of the dropped points for point dropping.

\begin{figure*}[htbp]
\begin{center}
  \includegraphics[width=0.98\linewidth]{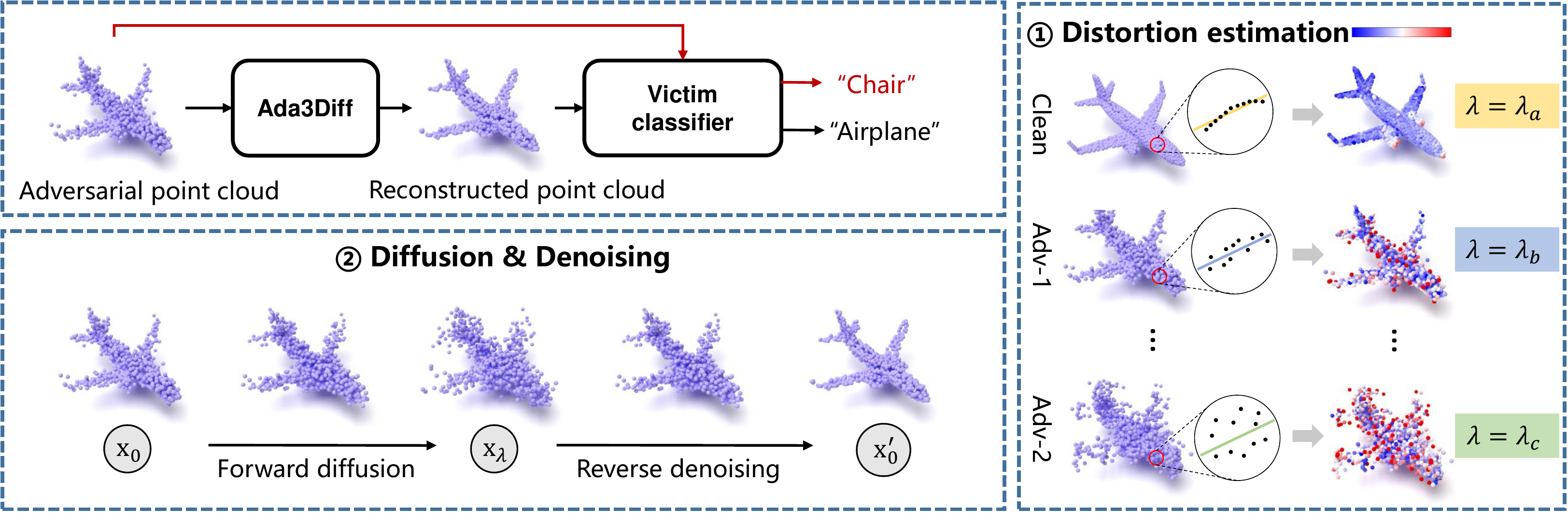}
\end{center}
  \caption{The overall framework of the proposed Ada3Diff, which consists of  \textcircled{1} distortion estimation and \textcircled{2} diffusion \& denoising. We estimate the distortion of the input point cloud $\mathbf{x}_0$ according to the distance of each point from its neighborhood best-fit plane where the color closer to red represents a higher degree of distortion. We adaptively choose different diffusion timesteps $\lambda$ ($\lambda_a<\lambda_b<\lambda_c$) for forward diffusion and then recover the clean distribution of the $\mathbf{x}_0$ by reverse denoising. }
\label{fig:overview}
\end{figure*}

Furthermore, we clarify the ability of attackers and our defense goal as follows:
1)
For defense methods integrated before the classifier input, when the attacker has access to all the information of the classifier without the knowledge of the preprocessor, it can be considered a gray-box attack. When the attacker has all the knowledge of the preprocessor and the classifier, then a white-box attack can be performed. 
2) As shown in Figure \ref{fig:overview},
given an adversarial point cloud  $\mathbf{x}^{adv}=\mathcal{A}(\mathbf{x})$,  Ada3Diff aims to purify it back to the clean distribution for correct prediction.

\subsection{Motivation}
Currently well-performing preprocessing-based defenses focus on using spatial priors to recover clean manifolds by first removing outliers from noisy point clouds and then optimizing the point distribution to approximate clean shapes. DUP-Net obtains dense and uniform point clouds with the help of a pre-trained upsampling network, and IF-Defense uses the implicit field pre-trained on clean point clouds and repulsive loss to constrain the points to be uniformly distributed on the surface.
However, although these optimizations make the purified point cloud dense and uniform, satisfying certain properties of the clean shapes does not necessarily lead to correct classification.
To this end, We visualize the high-level feature distributions of the point clouds that were purified by DUP-Net and IF-Defense in Figure \ref{tsne:motivation}. It can be seen that there is a severe overlap between the data distributions of different categories, which deviates significantly from the original clean data distribution.
This indicates that limited spatial prior knowledge is insufficient to recover the adversarial distribution accurately. In addition to the dense and uniform properties, the relative relationships and reasonable distributions between different points are crucial for classification. However, existing defenses are ineffective in understanding the deep features of clean manifolds when facing stronger attacks, such as reconstructing a vase with double surfaces.
Therefore, \textit{how to find a spatial prior that embodies all properties of the clean manifold?}  Dealing with this issue is challenging because many abstract properties are hard to characterize.
Fortunately, the data distribution implicitly contains multi-level semantic relationships of the point cloud, so we directly restore the adversarial data distribution instead of optimizing the spatial prior.

Specifically, we utilize an unconditional diffusion model to model the data distribution of clean point clouds, align the adversarial point cloud with the diffusion process and seamlessly integrate the recovery of the adversarial distribution into the reverse denoising process.
Now we must consider another important question: \textit{how can we balance the fidelity and robustness of denoising? }
As demonstrated in DiffPure~\cite{DiffPure}, there exists a fixed diffusion timestep that preserves the image structure while effectively removing noise. 
However, different types of adversarial point clouds have varying degrees of distortion, and a fixed diffusion timestep cannot balance their restoration effectively.
For imperceptible adversarial point clouds, smaller timesteps are appropriate since the defense mainly adjusts the spatial distribution to retain local structural details. For adversarial point clouds with greater distortion, larger timesteps should be used to ensure robustness in worse cases.
Thus, in order to ensure robustness against unknown stronger attacks while minimizing the impact on adversarial point clouds with minor distortion, we propose the design of a distortion-aware defense framework to select the appropriate diffusion path for varying input.

\subsection{Preliminary}
\paragraph{Formulation of diffusion models.}
Given a point cloud $\mathbf{x} \in \mathbb{R}^{N \times 3}$ sampled from a real-data distribution $q(\mathbf{x})$, the diffusion process consists of a Markov process can be parameterized as Gaussian form: 
\begin{equation} \label{eq:forward1}
\begin{aligned}
q(\mathbf{x}_t|\mathbf{x}_{t-1})=\mathcal N(\mathbf{x}_t;\sqrt{1-\beta_t}\mathbf{x}_{t-1},\beta_t\mathbf{I}),
\end{aligned}
\end{equation}
where $\beta_t \in [0,1]$ for $t \in [1,T]$ is a constant that controls the noise step size. From the good properties of Markov and Gaussian distributions, the diffusion expression of $\mathbf{x}_t$  at the $t$ step can be easily obtained from $\mathbf{x}_0$:
\begin{equation} \label{eq:forward2}
\begin{aligned}
\mathbf{x}_t=\sqrt{\overline{\alpha}_t}\mathbf{x}_0+\sqrt{1-\overline{\alpha}_t}\epsilon,\epsilon \sim \mathcal{N}(0,\mathbf{I}),
\end{aligned}
\end{equation}
where $\alpha_t=1-\beta_t$ and $\overline{\alpha}_t=\textstyle\prod_{i=1}^{t}\alpha_i$.
The reverse process then learns the stepwise denoising through a neural network $p_{\theta}$, again defined as a Markov process with Gaussian transitions:
\begin{equation} \label{eq:reverse}
\begin{aligned}
p_{\theta}(\mathbf{x}_{0:T})&=p(\mathbf{x}_T)\textstyle\prod_{t=1}^{T}p_{\theta}(\mathbf{x}_{t-1}|\mathbf{x}_t), \\
p_{\theta}(\mathbf{x}_{t-1}|\mathbf{x}_t)&=\mathcal{N}(\mathbf{x}_{t-1};\mu_{\theta}(\mathbf{x}_t,t),\sigma^2_t\mathbf{I}),
\end{aligned}
\end{equation}
where the variance $\sigma^2_t$ can be left out of the training. After training, randomly sampled Gaussian data points can be gradually generated into a real point cloud by $p_{\theta}$.

\subsection{Distortion-aware Adaptive Diffusion}
The overall framework of the proposed method is shown in Figure \ref{fig:overview}, which consists of two main stages, namely distortion estimation, and diffusion $\&$ denoising. We will elaborate on them in detail.
\paragraph{Distortion estimation.} Imperceptible adversarial points generally move along the surface, while perturbations with large distortions exhibit the opposite behavior and have a larger average distance from the origin point set, as illustrated in part \textcircled{1} in Fig. \ref{fig:overview}. This chaos is specifically characterized by a more uniform distribution of local point sets within the sphere.  
Inspired by the observation, we approximate the disorder degree of the local region by fitting the best plane of the local neighborhood for each point and calculating the distance from that point to the plane.

Mathematically, define $\mathbf{x}_{sub}=\{x_i\} \cup \mathcal{N}(x_i) \in \mathbb{R}^{K \times 3}$ is a subset containing $K$ points from $\mathbf{x}$, where $x_i \in \mathbb{R}^3$ is the center point and the other points $x_j \in \mathcal{N}(x_i)$ are the non-repeating $K-1$ nearest neighbors of $x_i$. 
Our goal is to find an optimal plane for fitting the neighborhood points such that the sum of the distances from the points in $\mathbf{x}_{sub}$ to this plane is minimized. Formally, a plane $\mathbf{p}$ is determined by its normal vector $\mathbf{n}=(n_x,n_y,n_z)^T$ and a point $p= (p_x,p_y,p_z)$ on the plane:
\begin{equation} \label{eq:plane}
\begin{aligned}
\mathbf{p}: (x - p)\cdot\mathbf{n}=0.
\end{aligned}
\end{equation}
The distance $d$ of a point $x$ that is not on the plane to the $\mathbf{p}$ is:
\begin{equation} \label{eq:distance}
\begin{aligned}
d = \left|(x - p)\cdot\mathbf{n}\right|/\left| \mathbf{n}\right|.
\end{aligned}
\end{equation}
The normal vector of the fitted plane can be obtained by minimizing the objective $L_d$ as follows:
\begin{equation} \label{eq:ld}
\begin{aligned}
L_d &= \textstyle\sum_{x_j \in \mathbf{x}_{sub}} \left|(x_j-p)\cdot\mathbf{n}\right| 
=\|\mathbf{x}_{sub}' \cdot \mathbf{n} \|, \;
s.t. \left\|\mathbf{n}\right\|_2=1, 
\end{aligned}
\end{equation}
where $\mathbf{x}_{sub}'$ is the matrix that consists of each point of $\mathbf{x}_{sub}$ minus the point $p$.
By calculating the partial derivative of $p$, we can conclude that $p=\frac1{\left|x_{sub}\right|}\sum_{x \in x_{sub}} x_j$ is the center of mass of the point in the neighborhood.
After determining $\mathbf{x}_{sub}'$, we perform the singular decomposition on $\mathbf{x}_{sub}'$, \ie, $\mathbf{x}_{sub}'=U\Sigma V^H$. The last column of matrix V is the normal vector of the plane to be fitted. 
Based on this, we can calculate the distance from the current point $x_i$ to the plane by $d_i = \left|(x_i-p)\cdot \mathbf{n}\right|$ or the sum of the distances from all points in the $\mathbf{x}_{sub}$ to the plane as an estimate of the distortion of the point $x_i$. Finally, we take the mean of distortion estimates of all points denoted as $\mathcal{E}_{\mathbf{x}}={\textstyle\frac1N}{\textstyle\sum_{i=1}^N}d_i$, as an approximation to the degree of perturbation of the whole point cloud $\mathbf{x}$.
\begin{table*}[t] 
\caption{Comparison of robust classification accuracy ($\%$) on ModelNet40 under different attacks.} 
\label{tab:01}
\centering
\renewcommand{\arraystretch}{1}
\setlength\tabcolsep{3.4mm}{
\begin{tabular}{l|c|ccccccc}
\toprule[1pt]
         Defenses & Clean & 3D-Adv    & $k$NN-Adv & PGD  & AdvPC & AOF & Add-Adv & Drop-200 \\ \midrule[1pt]
CD ($\times 10^{-4}$)    &  0.00 &  0.59  & 1.20  & 6.35 & 17.2 & 21.3  &0.29 & 11.0 \\ 
No defense & \textbf{92.3}  & 0.00     & 7.21   & 0.00     & 0.85  & 0.00   & 0.00  & 75.2 \\ 
SRS \cite{srs}    & 86.8  & 84.9       & 82.6       & 74.8     & 26.8      & 18.2   & 86.5 & 57.7     \\
SOR~\cite{sor}      & 91.0  & \textbf{88.1}      & 78.2        & 64.0     & 16.9      & 8.27  & 86.8  & \textbf{78.8}   \\ 
DUP-Net~\cite{dupnet}    & 88.8  & 88.0 & 85.3   & 79.5 & 49.4  & 31.6 & 87.2 & 74.2 \\ 
IF-Defense~\cite{if-defense} & 86.7  & 86.6  & 85.7   & 83.7   & 62.3  & 47.3 &86.2 & 77.6 \\ 
DiffPure-3D~\cite{DiffPure} &86.4   &  85.9 & 86.0 & 86.5 & 85.4 & 83.6 & 85.4 &75.4 \\ \midrule
Ada3Diff   & 88.4 & 87.7      & \textbf{87.2}       & \textbf{87.6} & \textbf{85.9}  & \textbf{85.4} &\textbf{88.5}  & 78.1     \\
\bottomrule[1pt]
\end{tabular}}
\end{table*}

\begin{figure}[htbp]
\begin{center}
  \includegraphics[width=0.96\linewidth]{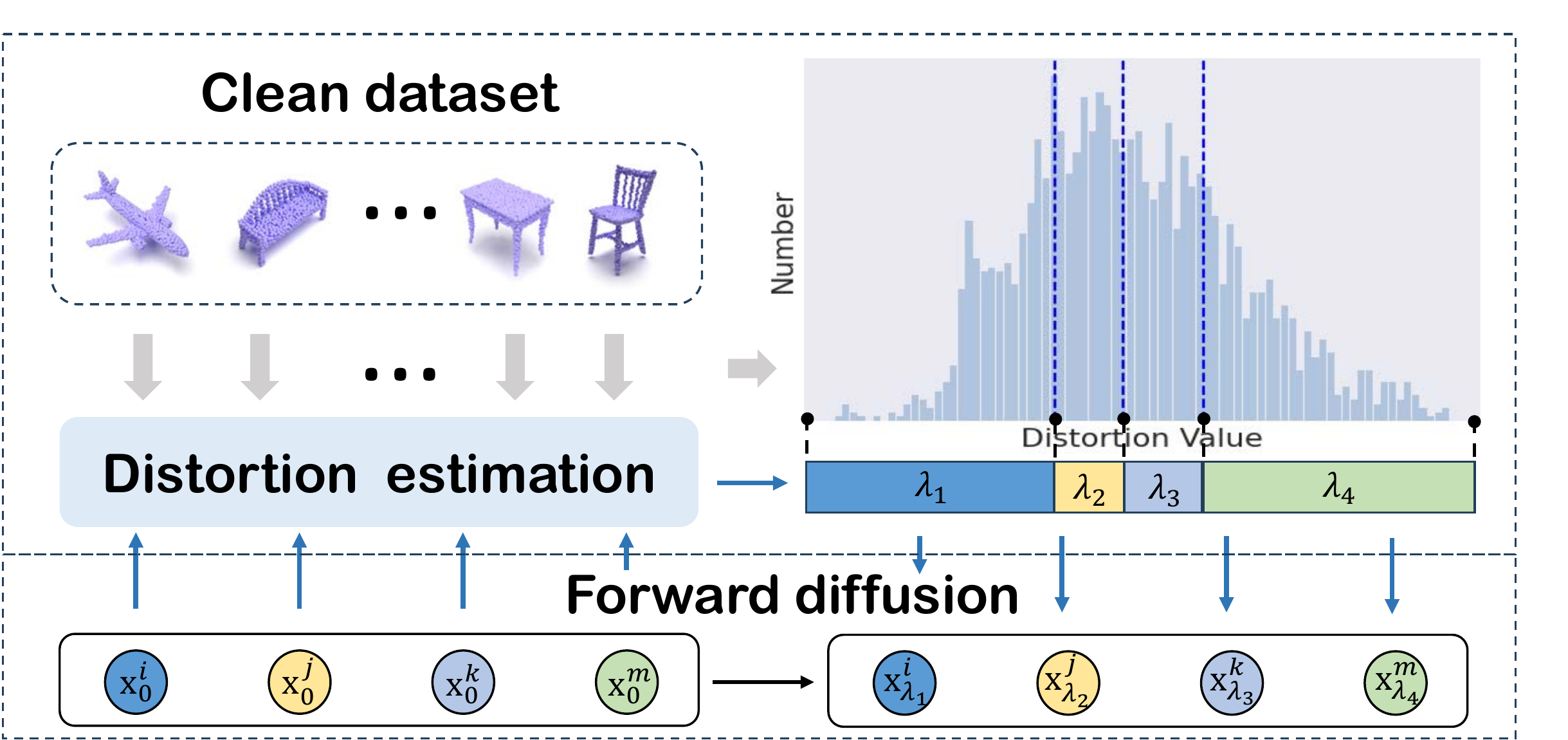}
\end{center}
  \caption{The visualization of adaptive timestep selection. Distortion thresholds are calculated by grouping the distribution of distortion values of the clean dataset.}
\label{fig:timestep}
\end{figure}

\paragraph{Diffusion and denoising.}
 Our diffusion-based denoising can be divided into two steps, which correspond to the adaptive forward diffusion process and the reverse denoising process. Given an adversarial point cloud $\mathbf{x}^{adv}_0 \in \mathbb{R}^{N \times 3}$, we first perform forward diffusion on it by ${\lambda}$ timesteps according to Eq.~(\ref{eq:forward2}):
\begin{equation} \label{eq:for3}
\begin{aligned}
\mathbf{x}^{adv}_{{\lambda}}
&=\sqrt{\overline{\alpha}_{\lambda}}\mathbf{x}^{adv}_0+\sqrt{1-\overline{\alpha}_{\lambda}}\epsilon \\
&=\sqrt{\overline{\alpha}_{\lambda}}\mathbf{x}_0+\sqrt{\overline{\alpha}_{\lambda}}\delta+\sqrt{1-\overline{\alpha}_{\lambda}}\epsilon,
\end{aligned}
\end{equation}
where $\delta$ satisfies a certain perturbation budget $\mathbf{B}_{eps}$. Each point in the point cloud is weighted with a superposition of independent random Gaussian noise, which counteracts the effect of the perturbation to some extent, as demonstrated in \cite{srs}. 
Furthermore, the forward diffusion is also used to make the starting point of the reverse process as consistent as possible with the input distribution of the Markov kernel at step ${\lambda}$. After that, we take $\mathbf{x}^{adv}_{{\lambda}}$ as the starting point and progressively eliminate the noise by $p_{\theta}(\mathbf{x}_{t-1}|\mathbf{x}_t)$. Finally, we output a reconstructed point cloud $\mathbf{x}'_{0}$ that is close to the clean data distribution and then feed it to the classifier.

As $\lambda$ increases, the point cloud gradually becomes noisy and the local structural details become increasingly rough due to the decreasing weight $\sqrt{\overline{\alpha}_{\lambda}}$ and the gradually increasing proportion of Gaussian noise. However, this helps to destroy the greater distortion to recover the adversarial point cloud with stronger attack performance. Therefore, the maximum timestep should be as small as possible while improving the robustness of the worse case.
Specifically, we choose the $\lambda$ that can exactly make the weighted Gaussian noise $\sqrt{1-\overline{\alpha}_{\lambda}}\epsilon$ larger than the weighted predetermined budget as the maximum diffusion timestep.

Figure \ref{fig:timestep} illustrates how the adaptive timestep is selected. 
Specifically, we first calculate $\mathcal{E}_{\mathbf{x}}$ of all clean point clouds in the dataset and get the distortion distribution $\mathcal{D}$ of the whole clean dataset. Then, we quarter the distribution $\mathcal{D}$ as $\mathcal{D}_{1}$, $\mathcal{D}_{2}$, $\mathcal{D}_{3}$, and $\mathcal{D}_{4}$ in turn. Similarly, we quarter the maximum diffusion timestep $\lambda$ as $\lambda_1$, $\lambda_2$, $\lambda_3$, and $\lambda_4$, respectively. Finally, given a target point cloud $\mathbf{x}_i$, we select the corresponding timestep $\lambda_i$ based on the distortion range $\mathcal{D}_{i}$, which the $\mathcal{E}_{\mathbf{x_i}}$ belongs to.
The above strategy is reasonable because the clean point cloud with a large local curvature or the presence of multiple surfaces in relatively close proximity can also make the $\mathcal{E}_{\mathbf{x_i}}$ large. The distortion values estimated with different-size datasets are similar and the corresponding analysis is provided in the supplementary material.
By adaptive diffusion and denoising, we expect to preserve more local structures for point clouds with good surface properties and improve their classification accuracy.

\section{Experiment}
\subsection{Experimental Setup}
\paragraph{Datasets and recognition models.}
We adopt ModelNet40~\cite{ModelNet} as the default dataset for comparison and main experiments. Besides, we also leverage ShapeNet  Part~\cite{ShapeNetPart} dataset to verify the generality of Ada3Diff. 
For point cloud recognition models, we utilize CurveNet~\cite{curvenet} for comparison and demonstrate generality by other classic architectures including PointNet~\cite{pn} based on MLP, DGCNN~\cite{dgcnn} based on graph convolution, PCT~\cite{pct} using a transformer, and PointConv~\cite{pointconv} based on convolution.
More details about datasets can be found in the appendix.
\paragraph{Adversarial attacks and the baseline defenses.}
We selected attack methods with different imperceptibility, namely 3D-Adv~\cite{cw}, $k$NN-Adv~\cite{cw-knn}, PGD~\cite{pgd,xiaoyi}, AdvPC~\cite{advpc},  AOF~\cite{aof}, Add-Adv~\cite{cw} and point dropping~\cite{drop}. 
We set the maximum perturbation distance of all attacks except point dropping and Add-Adv to $l_{\infty}$ = 0.18. 

We compare the proposed method with the baseline defense methods that also target pre-processing the adversarial point cloud before recognition, including SRS~\cite{srs}, SOR~\cite{sor}, DUP-Net~\cite{dupnet} and IF-Defense~\cite{if-defense}, where SOR we refer to the implementation in DUP-Net.  
We follow the default settings in their original paper or released code. 
We also extend the DiffPure~\cite{DiffPure} strategy to 3D point clouds for comparison in the main results.
More details about attacks and the baselines can be found in the supplementary materials. 

\begin{table*}[t] 
\caption{Robustness evaluation ($\%$) on ShapeNet Part under different attacks.}
\label{tab:02}
\centering
\renewcommand{\arraystretch}{1}
\setlength\tabcolsep{3.2mm}{
\begin{tabular}{l|c|ccccccc}
\toprule[1pt]
        Defenses  & Clean & 3D-Adv    & $k$NN-Adv & PGD  & AdvPC & AOF  & Add-Adv & Drop-200 \\ \midrule[1pt]
No defense & 98.5  & 0.00     & 16.6   & 8.94     & 15.0  & 0.14  &0.00 & 87.9 \\ 
Ada3Diff   & 98.2 & 97.9      & 98.0     & 98.0 & 97.4  & 96.8 &98.4 & 91.7     \\
\bottomrule[1pt]
\end{tabular}}
\end{table*}

\begin{table*}[t] 
\caption{Robustness evaluation ($\%$) against various attacks on different models. All experiments are conducted on ModelNet40.}
\label{tab:03}
\centering
\renewcommand{\arraystretch}{1}
\begin{tabular}{l|c|cccccccc}
\toprule[1pt]
Models         & NaturalAcc & Clean & 3D-Adv    & $k$NN-Adv & PGD  & AdvPC & AOF &Add-Adv & Drop-200 \\ \midrule[1pt]
PointNet~\cite{pn}        & 89.0 & 86.9 & 86.9       & 86.2       & 85.9     & 85.0      & 84.1     &86.8 & 55.1     \\ 
DGCNN~\cite{dgcnn}        & 92.3 & 87.2 & 86.7      & 86.4        & 85.9     & 85.9      & 85.3     &86.6 & 75.6   \\ 
PCT~\cite{pct}    & 92.7 & 87.6 & 87.6 & 86.9   & 86.5 & 85.5  & 85.8 &87.3 & 71.8 \\ 
PointConv~\cite{pointconv} & 91.9 & 86.8 & 85.5  & 85.6   & 86.3   & 83.4  & 83.5 &87.0 & 68.9 \\ 
CurveNet~\cite{curvenet}   & 92.3  & 88.4 & 87.7      & 87.2       & 87.6 & 85.9  & 85.4 &88.5 & 78.1     \\
\bottomrule[1pt]
\end{tabular}
\end{table*}

\paragraph{Implementation details.}
Following PVD~\cite{pvd}, we use point-voxel CNN~\cite{pvcnn} as the backbone of the diffusion model and train it on ModelNet40 and ShapeNet Part, respectively. 
We select 10 nearest neighbors around each point to estimate the distortion. 
We set $\lambda=20$ as the maximum diffusion timestep to improve white-box robustness, which is equivalent to choosing a 98.8$\%$ confidence interval for Gaussian distribution in Eq. \ref{eq:for3}.
We empirically operate 4 rounds to obtain better robustness against adaptive attacks.

\paragraph{Evaluation metrics.}
In this paper, we mainly focus on the robustness evaluation. For this, we adopt robust classification accuracy, which is defined as the accuracy of the set of samples that are purified by defense methods. 
We use the average Chamfer distance (CD)~\cite{chamfer} between the perturbation-based adversarial point cloud and the original point cloud to measure the perturbation strength. Higher CD means larger distortions.

\subsection{Comparison with Baseline Defense Methods}

\paragraph{Robustness against various attacks.}
In Table \ref{tab:01}, we provide the comparison of the robust classification accuracy with the baseline defense methods under different adversarial attacks. It can be seen that the proposed Ada3Diff performs well in most cases and outperforms other defense methods in most cases by a large margin especially suffering adversarial attacks with severe distortion, such as AdvPC and AOF.
The proposed method can achieve slightly superior performance to the best defense for attacks with slight distortions, such as 3D-Adv, kNN-Adv, and PGD.
For the special attack Drop-200 that removes 200 crucial points based on the saliency map, all defense methods perform poorly, where our performance is still acceptable. The defense is unable to effectively recover the deleted points due to the constraints of the existing diffusion timestep budget. We suggest using a larger diffusion timestep and adding additional priors to guide the denoising process to solve this problem.
Overall, although the present 3D point cloud defense methods can effectively defend against common shape-invariant attacks, it is still vulnerable to stronger attacks, while our method shows comprehensive robustness by recovering the data distribution in the face of different distorted adversarial point clouds.

\begin{figure*}[htbp]
\begin{center}
  \includegraphics[width=0.95\linewidth]{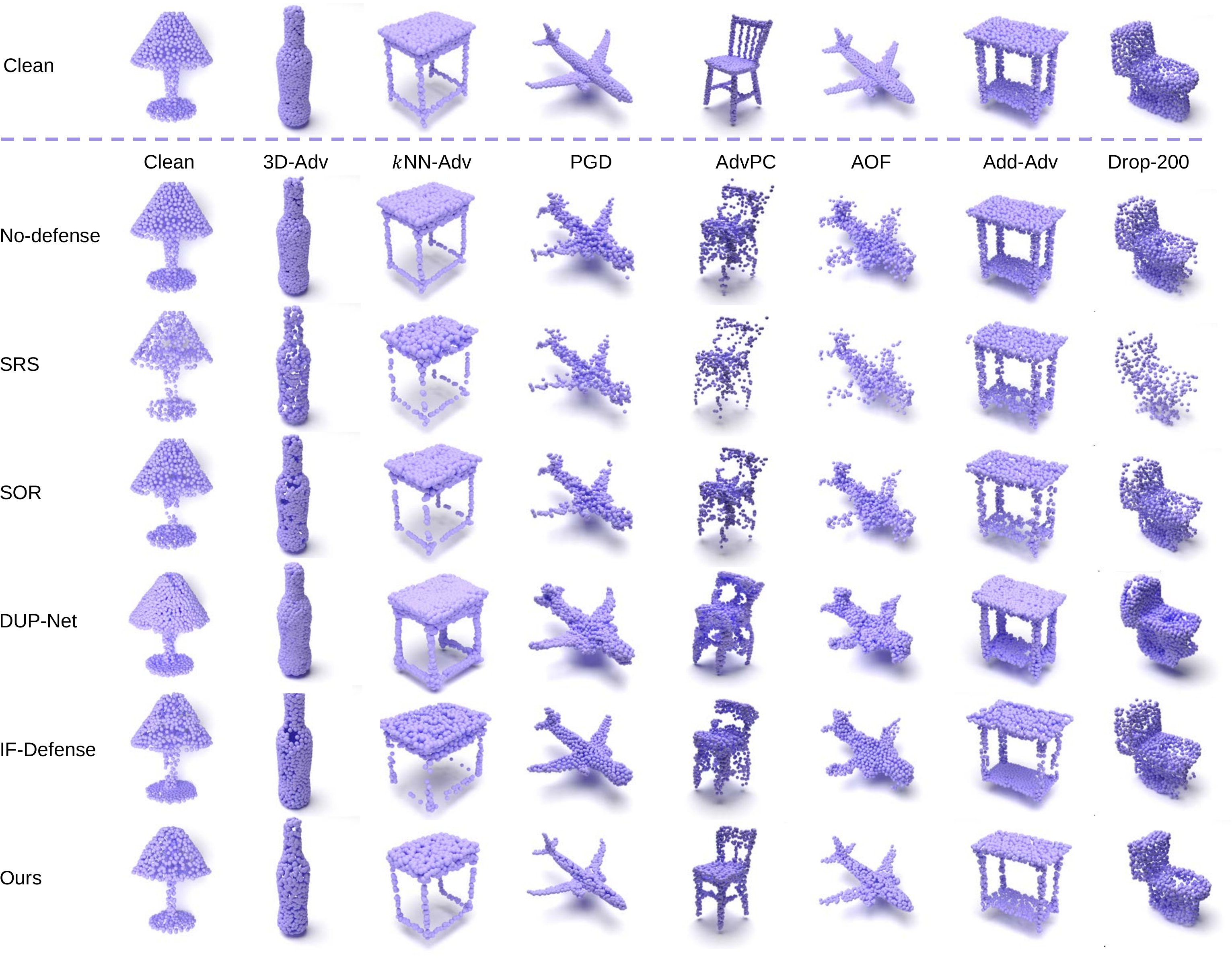}
\end{center}
  \caption{Qualitative comparison of reconstruction results of different defense methods against diverse adversarial point clouds. 
   The first row is the corresponding clean point clouds. Each column shows the reconstruction results for different defense methods. Note: The DiffPure visualization results are similar to ours and are provided in the Supplementary Material.
  } 
  \label{fig:visualization}
\end{figure*}

\paragraph{Point distribution after defenses.}
In Figure \ref{fig:visualization}, we provide some visual examples of the purified point clouds by different defenses. Consistent with the quantitative results in Table \ref{tab:01}, Ada3Diff achieves an excellent reconstruction suffering attacks with large distortions, while the point clouds processed by other methods are either cluttered or the points are not well distributed on the surface such as chairs and airplanes under AOF. Ada3Diff is able to reconstruct details better for adversarial point clouds with various distortions, such as lamp supports, table legs, and chair legs. 
As attacks become stronger and distortions increase, it becomes challenging to utilize the spatial manifold prior for point cloud reconstruction. Conversely, the data distribution prior, which is derived by modeling from noisy to real point clouds, demonstrates superior generalization capabilities across various noisy distributions.

\renewcommand{\dblfloatpagefraction}{0.9}
\subsection{Comprehensive Evaluation on Ada3Diff}
\paragraph{Evaluation on different datasets.}
We modeled the distribution of the ShapeNet Part dataset and generated adversarial point clouds using the same attack methods as ModelNet40.
Table \ref{tab:02} shows the robust classification accuracy of Ada3Diff on the ShapeNet Part dataset. 
The proposed approach still shows superb defense performance against different attacks.

\begin{figure*}[h!tb]
\centering
\subcaptionbox{}{
\begin{minipage}[t]{0.325\linewidth}
\centering
\includegraphics[width=\linewidth]{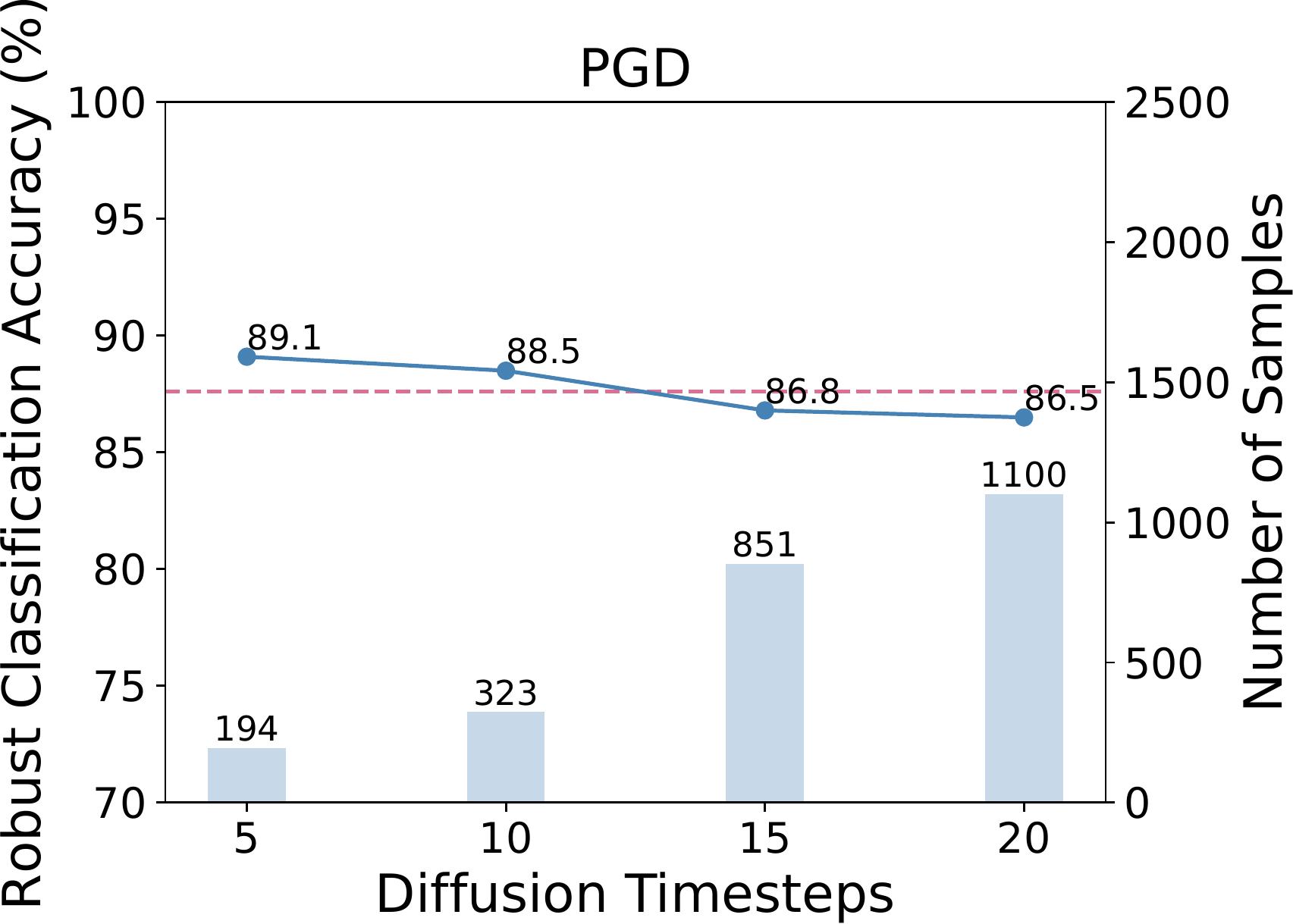}
\end{minipage}%
}%
\subcaptionbox{}{
\begin{minipage}[t]{0.32\linewidth}
\centering
\includegraphics[width=\linewidth]{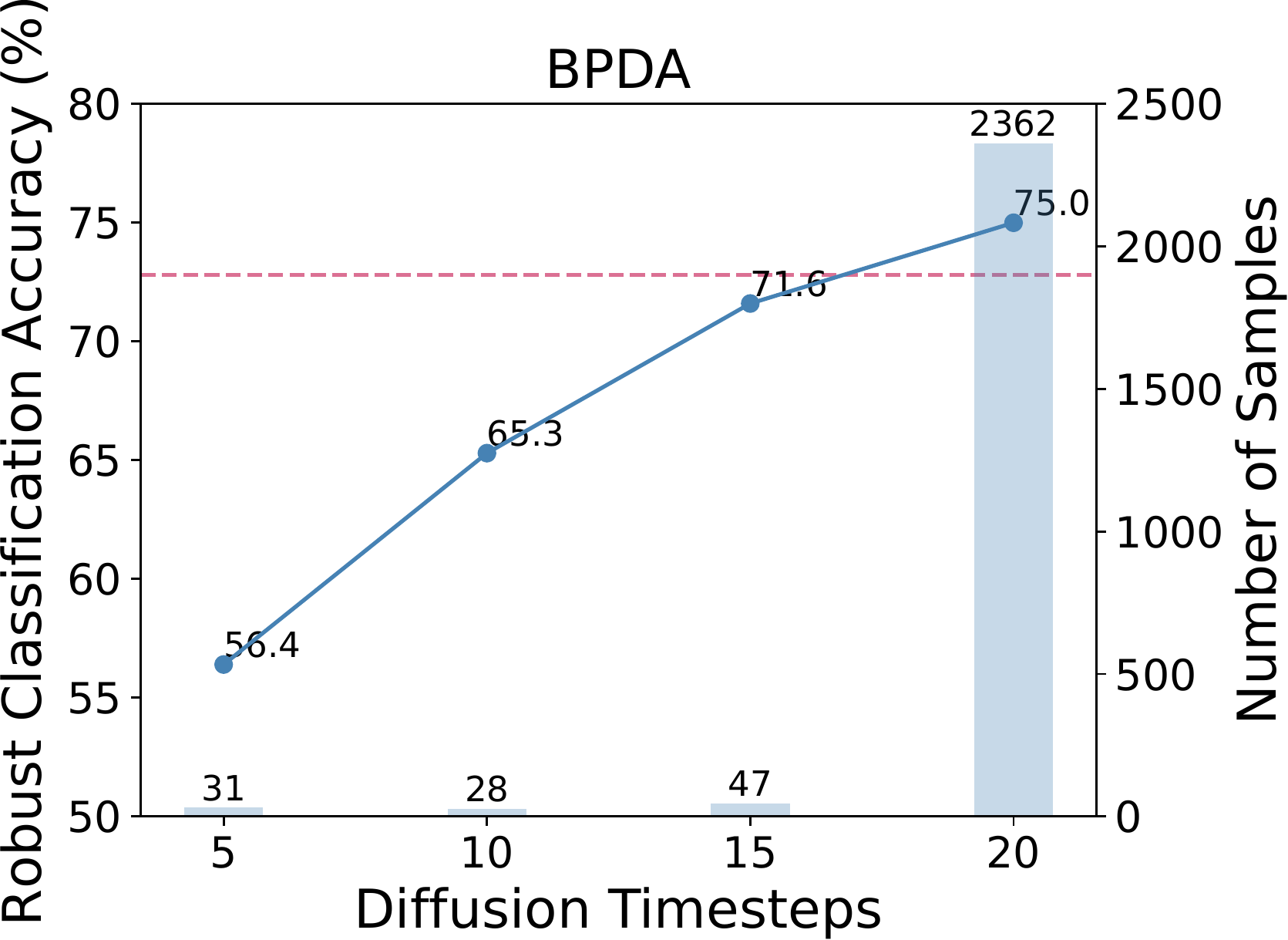}
\end{minipage}%
}%
\subcaptionbox{}{
\begin{minipage}[t]{0.277\linewidth}
\centering
\includegraphics[width=\linewidth]{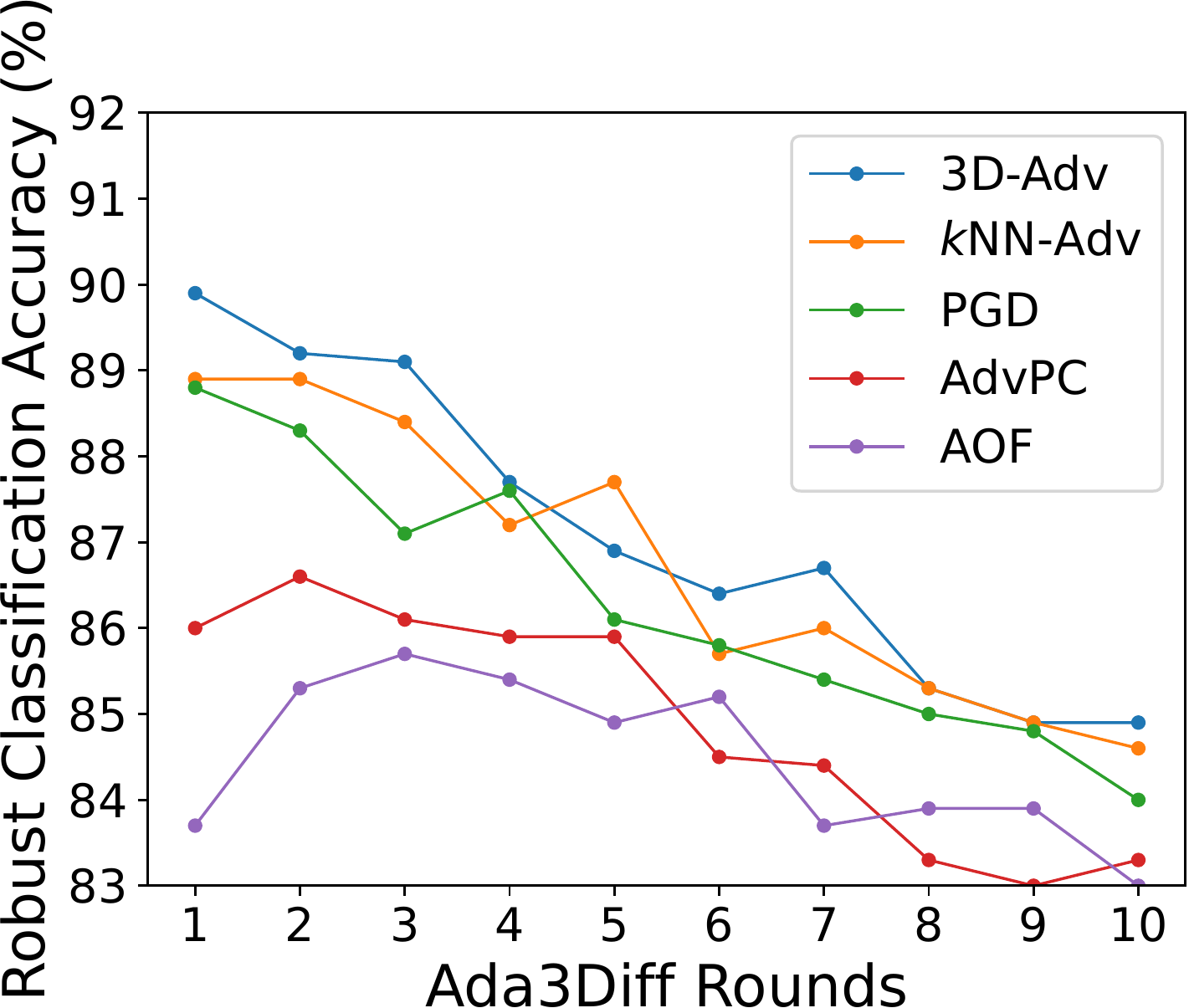}
\end{minipage}%
}%
\centering

\caption{The influence of the diffusion timesteps and Ada3Diff rounds. In (a) and (b): the blue line means the robustness with fixed diffusion timesteps, where the red dashed line represents the corresponding performance of Ada3Diff. We also show the sampling numbers of each timestep by Ada3Diff. (c) shows the performance of the defense with different rounds.}
\label{fig:ablation}
\end{figure*}
\paragraph{Evaluation on different models.}
To further evaluate the generality of the proposed approach to different models, we conducted experiments on point cloud recognition networks with different architectures. 
As shown in Table \ref{tab:03}, the robust classification accuracy under different attacks other than point dropping is basically above 85$\%$ for all recognition models. 
Because these recognition models focus on different spatial relationships, containing global relationships as well as local correlations from various local aggregation operations, it means that the recovered point cloud is well reconstructed in terms of both local structural features and global consistency. 
Therefore, Ada3Diff can be used as a plug-and-play pre-processing module that can be attached to any 3D point cloud recognition model to enhance robustness.

\paragraph{Evaluation on different attack scenarios.}

Previous studies~\cite{jgba,adaptive} have demonstrated that defenses such as SOR or DUP-Net would lose the original robustness when non-differentiable operations are incorporated into the attack pipeline. Therefore it is necessary to evaluate whether the defense provides solid robustness.
To evaluate the robustness of the proposed method against adaptive attacks, we incorporate  Ada3Diff into the attack pipeline and perform a Backward Pass Differentiable Approximation (BPDA)~\cite{bpda} based PGD attack in the white-box scenario. Assuming that the point cloud permutation after being purified is unchanged, we use BPDA to obtain the approximate gradient of the purified point cloud with respect to the input, so as to perform a PGD attack in an end-to-end manner. We increase the number of gradient descents to 100.  
Table\ref{tab:bpda} shows the robust classification accuracy of defense against the BPDA adaptive attack, and we find that Ada3Diff can still maintain a robust accuracy of 72.8$\%$, which is even better than the performance of other methods in the gray box case.

\begin{table}[t]
\caption{Robust classification accuracy ($\%$) against the BPDA-based white-box attack (for Ada3Diff) on ModelNet40.}
\label{tab:bpda}
\setlength\tabcolsep{2pt}
\centering
\begin{tabular}{cccccc}
\toprule[1pt]
        Attack         & SRS  & SOR  & DUP-Net & IF-Defense & Ada3Diff   \\ \midrule[1pt]
BPDA-PGD  & 12.6 & 15.5 & 22.6    & 32.0 & \textbf{72.8}     \\
\bottomrule[1pt]
\end{tabular}
\end{table}

\begin{figure}[htbp]
\centering
\subcaptionbox{PGD}{
\begin{minipage}[t]{0.46\linewidth}
\centering
\includegraphics[width=\linewidth]{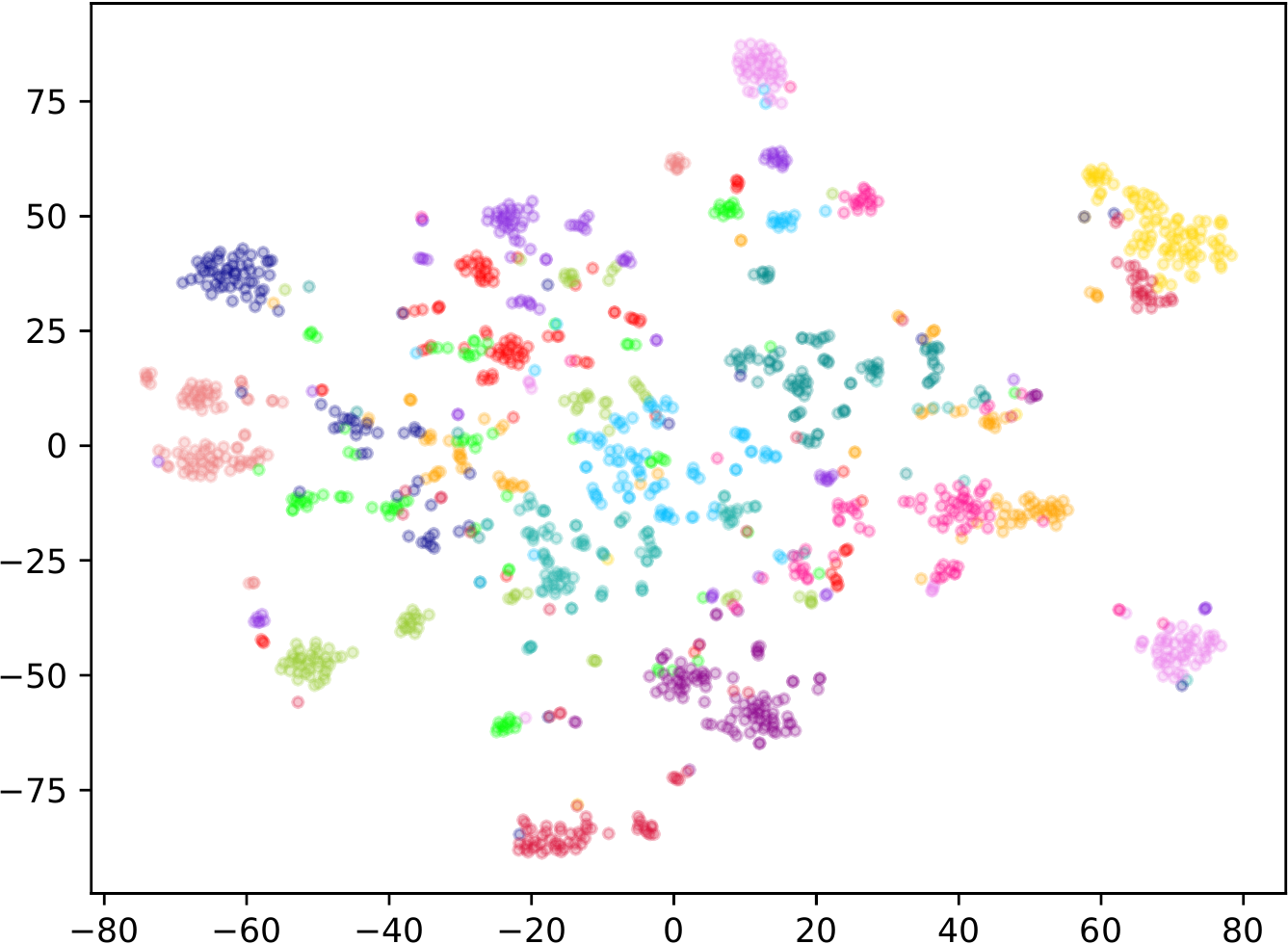}
\end{minipage}%
}%
\subcaptionbox{Ada3Diff-PGD}{
\begin{minipage}[t]{0.46\linewidth}
\centering
\includegraphics[width=\linewidth]{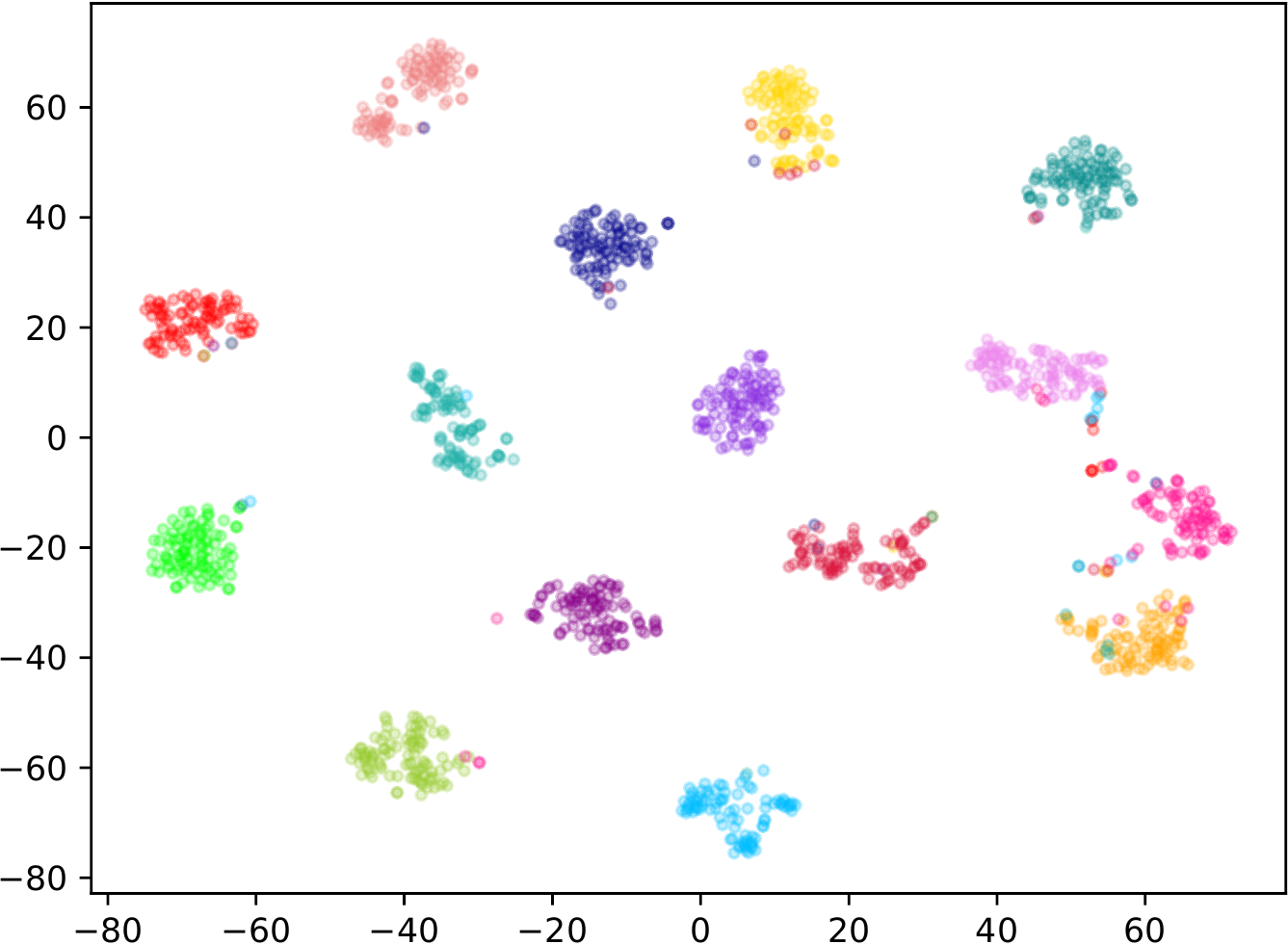}
\end{minipage}%
}%
\\
\subcaptionbox{AOF}{
\begin{minipage}[t]{0.46\linewidth}
\centering
\includegraphics[width=\linewidth]{figures/tsne-seed2023/aof/aof_linf_0.18-c.pdf}
\end{minipage}%
}%
\subcaptionbox{Ada3Diff-AOF}{
\begin{minipage}[t]{0.46\linewidth}
\centering
\includegraphics[width=\linewidth]{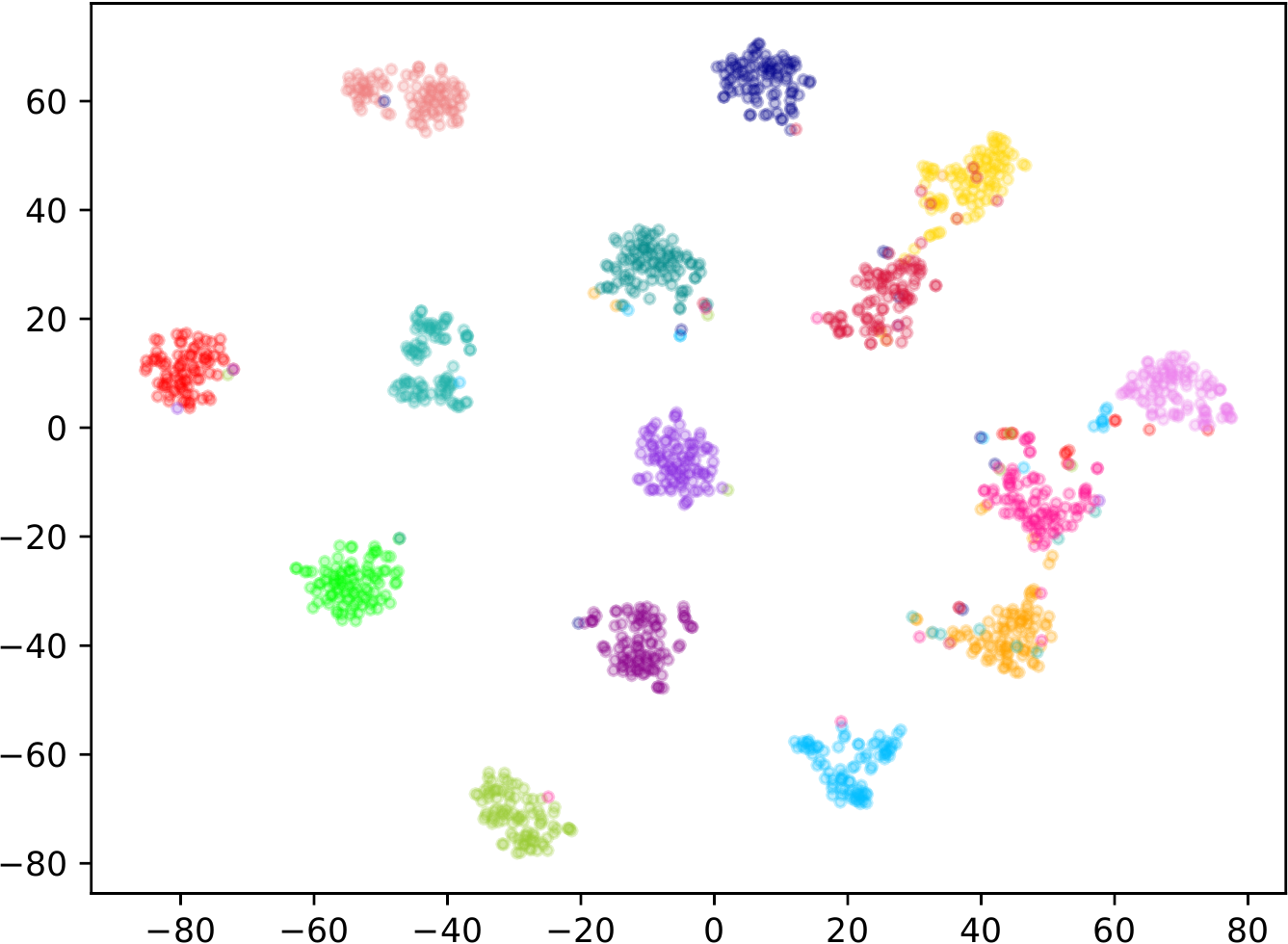}
\end{minipage}%
}%
\centering
\caption{The t-SNE visualization of latent distribution. The first column shows the adversarial distribution, and the second column shows the distribution after Ada3Diff recovery.
}
\label{tsne:01}
\end{figure}

\paragraph{Qualitative results of restored adversarial distribution.}
Figure~\ref{tsne:01} shows the adversarial distribution and the corresponding restored distribution.
The denoised point clouds can be correctly clustered with well-defined boundaries between categories. Moreover, the relative distributions of the categories are similar across different attacks and also similar to the clean distribution in Figure \ref{tsne:motivation} (a). 
Guided by the clean data distribution, Ada3Diff can recover adversarial distributions with greater distortions at acceptable offsets.
\paragraph{The influence of diffusion timesteps and rounds.}
In Figure \ref{fig:ablation}(a) and \ref{fig:ablation}(b), the red dashed lines demonstrate that the optimal diffusion timesteps are different for attacks with different distortions. Smaller timesteps help to achieve greater robust accuracy for adversarial point clouds with modest distortion. The proposed method can well balance the defense performance against various attacks. 
It ensures white-box robustness while minimizing the impact on the robust accuracy of the adversarial point cloud with small distortions.
Moreover, we compared Ada3Diff with the DiffPure~\cite{DiffPure} strategy at more maximum timesteps. Both use the same number of rounds.
The results are presented in Table \ref{tab:R1-Q1}. It can be seen that Ada3Diff using adaptive diffusion improves the robustness accuracy for different attacks at different time diffusion steps.
As the maximum time step increases, the diffusion range also becomes larger, and the accuracy improvement brought by Ada3Diff becomes more pronounced.
The Ada3Diff allows defenders to use larger diffusion timesteps to enhance the robustness against adaptive attacks.

We also considered different rounds to evaluate the effect of the number of iterations.
Figure \ref{fig:ablation}(c) shows that more rounds will induce degradation of robust accuracy, while fewer rounds cannot resist attacks with large distortions such as AOF. Furthermore, multi-round denoising can improve worst-case robustness for the adaptive attack. We find rounds from 2 to 4 have better defensive performance and can be chosen flexibly in practice.

\begin{table}[htbp] 
\caption{Comparison of robust classification accuracy ($\%$) under different timesteps. `DiffPure-3D' represents the version of DiffPure extended to 3D point clouds.}
\label{tab:R1-Q1}
\centering
\setlength{\tabcolsep}{2pt}
\begin{tabular}{c|cccccc}
\toprule[1pt]
         Defenses &Clean & 3D-Adv    & $k$NN-Adv & PGD  & AdvPC & AOF \\ \midrule[1pt]
DiffPure-3D (t=10)  & 89.1 &  88.6 & 88.3 & 88.5 & \textbf{86.4} & 85.6   \\ 
\rowcolor{gray!10}Ada3Diff (t=10) & \textbf{89.7} &  \textbf{89.5} & \textbf{88.6} & \textbf{88.5} & 86.3 & \textbf{85.7}   \\ \midrule
DiffPure-3D (t=15) & 87.2  &  87.4 & 87.3 & 86.8 & 86.0 & 85.8 \\ 
\rowcolor{gray!10}Ada3Diff (t=15) & \textbf{89.1} &  \textbf{88.8} & \textbf{87.4} & \textbf{87.2} & \textbf{86.4} & \textbf{85.9}    \\ \midrule
DiffPure-3D (t=20) &86.4   &  85.9 & 86.0 & 86.5 & 85.4 & 83.6\\ 
\rowcolor{gray!10}Ada3Diff (t=20) & \textbf{88.4}  & \textbf{87.7}    & \textbf{87.2}      & \textbf{87.6} & \textbf{85.9}  & \textbf{85.4}      \\
\bottomrule[1pt]
\end{tabular}
\end{table}
\section{Conclusion}
In this paper, we propose a novel 3D point cloud defense framework, named Ada3Diff, to reconstruct adversarial point clouds by distortion-aware adaptive diffusion. The proposed Ada3Diff can be plugged into various classification networks to help robust recognition. Extensive experiments have demonstrated that Ada3Diff can effectively defend against point cloud attacks with various distortions. We outperform present defenses by a large margin especially suffering large distortions. We shall point out that the proposed distortion estimation and diffusion step interval division can be further studied to improve defensive performance. Besides, how to enhance the robustness against the point-dropping attack is also a research direction for our future work.

\begin{acks}
This work was supported in part by the National Natural Science Foundation of China under Grant U20B2047, 62072421, 62002334, 62102386, and 62121002.
\end{acks}

\bibliographystyle{ACM-Reference-Format}
\balance
\bibliography{sample-sigconf}

\clearpage
\appendix
% \section*{\centering Appendix}

\twocolumn[
\begin{center}
% {\Huge\bfseries Ada3Diff: Defending against 3D Adversarial Point Clouds via
% Adaptive Diffusion}\\
% \vspace{0.2cm}
{\Huge\bfseries Supplementary Material}
\vspace{0.5cm}
\end{center}
]

% \titleformat*{\section}{\normalfont\Large\bfseries}
% \section*{Supplementary Material}
% \titleformatmy*{\section}{\normalfont\Large\bfseries}

\section{Details of Experiment Setup}

\paragraph{Datasets} 
We conducted experiments on two datasets, ModelNet40 and ShapeNet Part. ModelNet40 contains 12,311 CAD models with 40 semantic categories, of which 2,468 are for testing. ShapeNet Part has 16 categories with 16,846 objects, of which 2,874 are for testing. For both datasets, we sample 1,024 points uniformly from the surface of each sample and normalize them to a unit cube. 

\paragraph{Attacks}
We mainly selected six attack methods for generating adversarial point clouds with different degrees of distortion and additionally used point-dropping attacks beyond distortion measures.
For 3D-Adv~\cite{cw}, we apply $L_2$ distance for perturbation constraint, using 10 binary searches with 200 iterations each. The weight of the L2 loss is initially 10 and a maximum of 80. 
Add-Adv~\cite{cw} is a version of 3D-Adv with added points. We use chamfer distance and add 512 points for untargeted attacks. 
$k$NN-Adv~\cite{cw-knn} employs Chamfer measurement and $k$-NN loss, with weights 5 and 3 respectively. Optimization involves 1000 iterations using Adam, starting the learning rate at 0.001.
For PGD~\cite{pgd} attack, we use 50 iterations and a step size of 0.007 as in \cite{siadv}, employing the max-margin loss instead of cross-entropy loss.
AdvPC~\cite{advpc} is a transferability-based attack, we use $\gamma$ = 0.25 and iterate 1000 times.
AOF~\cite{aof} combines the low-frequency component in the adversarial loss used to improve the adversarial transferability. We follow the default settings in the paper. 
We set 200 points to be discarded for the point dropping~\cite{drop} attack, namely, Drop-200.
For the BPDA attack, the gradient is still an approximate value even after many iterations, so it is not the optimal adaptive attack and can still be improved in future work.

\paragraph{Defenses}
Given a point cloud with $N$ points, we randomly sample $N-500$ points for SRS~\cite{srs} defense. For SOR~\cite{sor}, we follow the default settings in \cite{dupnet}. We set the upsampling ratio to 4 for DUP-Net~\cite{dupnet}, i.e. 4096 points for the final output. IF-Defense~\cite{if-defense} utilizes the prior of implicit functions and has the best defensive performance among present pre-processing methods. We optimize 200 times of this defense based on ConvONet~\cite{convnet}, following its setting. We also extend the DiffPure~\cite{DiffPure} strategy to 3D point cloud defense. Specifically, DiffPure uses a fixed diffusion timestep in the experiments, and other settings are the same as Ada3Diff.

\section{Experiment and Discussion}
\subsection{More results}
\begin{table*}[htbp]
\centering
\caption{Comparison of robust classification accuracy ($\%$) of PointNet on ModelNet40 under different attacks.}
\label{tab:06}
\renewcommand{\arraystretch}{1}
{
\begin{tabular}{cl|c|ccccccc|c}
\toprule[1pt]
          Model &Defense & Clean & 3D-Adv    & $k$NN-Adv & PGD  & AdvPC & AOF  & Add-Adv & Drop-200 & Avg     \\ \midrule[1pt]
\multirow{6}{*}{PointNet} &No defense & 89.0  & 0.00              & 6.44          & 0.00              & 0.00            & 0.00             & 0.00 & 38.9 & 6.47              \\ 
&SRS        & 89.3  & 87.6          & 82.6          & 72.0          & 7.78            & 1.70        & 85.0  & 42.1   & 54.1\\ 
&SOR        & \textbf{89.4}  & \textbf{88.2} & 80.1          & 78.8          & 47.5            & 5.39        &83.3  & 42.7      & 60.9    \\ 
&DUP-Net    & 88.1  & 87.8          & 84.1          & 83.1          & 63.3        & 14.3      & \textbf{87.1}   & 46.1        & 66.5  \\ 
&IF-Defense & 87.4  & 87.3          & \textbf{86.5} & \textbf{86.9}          & 81.5        & 54.8       & 87.0  & \textbf{55.5} & 77.1\\ \cmidrule{2-11}
&Ada3Diff   & 89.0  & 86.9          & 86.2          & 85.9 & \textbf{85.0} & \textbf{84.1} &87.0 & 55.1   & \textbf{81.5}      \\ \midrule[1pt]

\multirow{6}{*}{DGCNN} &No defense & \textbf{92.3}  & 0.00 & 2.80       & 0.00          & 2.10    &0.00    &0.00 & 69.1   &10.6  \\ 
&SRS        & 83.5  & 80.5 & 78     & 69.3      & 24.7  & 17.4 & \textbf{87.9} & 46.1 & 57.7 \\ 
&SOR        & 91.3  & \textbf{87.4} & 72.7   & 70.3       & 23.6  & 12.7 &73.4 & 70.9 & 58.7\\ 
&DUP-Net    & 64.0  & 59.4 & 46.8   & 44.6      & 20.1  & 13.3 &43.6 & 43.5 & 38.8\\ 
&IF-Defense & 84.2  & 84.4 & 84.0   & 83.1       & 70.3  & 51.7 & 81.4 & 72.7 & 75.4 \\ \cmidrule{2-11}
&Ada3Diff   & 87.2  & 86.7 & \textbf{86.4}   & \textbf{85.9}          & \textbf{85.9}  & \textbf{85.3} & 86.6 & \textbf{75.6}  &  \textbf{84.6} \\ \midrule[1pt]

\multirow{6}{*}{PCT} &No defense  & \textbf{92.7}  & 75.6              & 7.74          & 0.28          & 0.53          & 2.15     & 73.3 & 66.9      & 32.4   \\
&SRS         & 91.8  & 90.9          & 79.9          & 74.0      & 32.9      & 20.3 & 88.2   & 69.4   & 65.1  \\ 
&SOR         & 91.8  & \textbf{91.7} & 66.9          & 62.2      & 29.2      & 12.8  & \textbf{90.4}  & 71.3   & 60.6  \\ 
&DUP-Net     & 85.0  & 82.9          & 72.9          & 63.0      & 31.8      & 21.9  &81.2  & 59.0    & 59.0   \\ 
&IF-Defense  & 88.9  & 88.9          & 86.1 & 82.7      & 67.8      & 57.1 & 87.4  & 71.4  & 77.3   \\ \cmidrule{2-11}
&Ada3Diff    & 87.4  & 87.6             & \textbf{86.9} & \textbf{86.5} & \textbf{85.5} & \textbf{85.8}    & 87.7   & \textbf{71.8}     & \textbf{84.5}   \\ \midrule[1pt]

\multirow{6}{*}{PointConv} &No defense & \textbf{91.9} & 0.00             & 7.62          & 0.00             & 0.41          & 0.00          &0.00   & 78.0   & 12.3        \\ 
 &SRS        & 90.7          & 88.7          & 79.5          & 71.1          & 12.1          & 11.6       & \textbf{90.6}   & 71.2          & 60.7\\ 
 &SOR        & 91.3          & \textbf{90.4} & 71.8          & 49.8          & 3.90           & 5.31       &88.5   & \textbf{78.7} & 55.5\\ 
 &DUP-Net    & 84.1          & 82.2          & 78.2          & 69.2          & 32.9          & 33.3       & 81.6  & 68.6         & 63.7\\
 &IF-Defense & 89.7          & 89.2          & \textbf{88.0} & 85.0          & 46.6          & 43.6       & 89.4  & 76.9          & 74.1\\ \cmidrule{2-11}
 &Ada3Diff    & 86.8          & 86.9          & 86.5          & \textbf{86.3} & \textbf{83.4} & \textbf{83.5} & 87.1 & 69.7  & \textbf{83.3} \\ \bottomrule[1pt]      
\end{tabular}}
\end{table*}

\begin{figure}[htbp]
\begin{center}
  \includegraphics[width=1\linewidth]{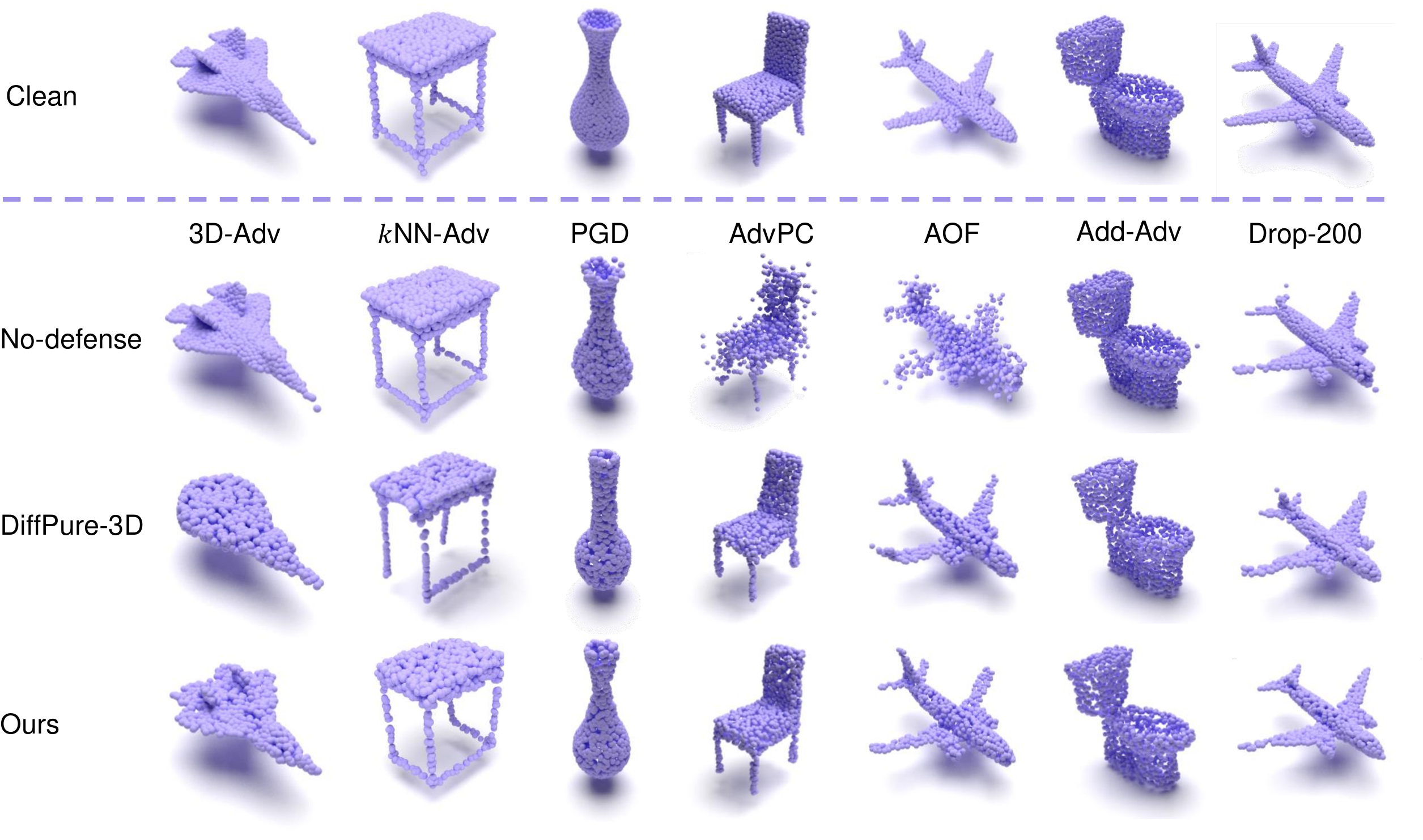}
\end{center}
  \caption{Visualization of the reconstructed point cloud.} 
  \label{compare}
\end{figure}

\paragraph{Visualization of reconstructed point clouds.}
Figure \ref{compare} shows visual comparison between DiffPure-3D and Ada3Diff. Point clouds reconstructed from adversarial samples by our method are slightly better with more shape details preserved.

\begin{figure}[htbp]
\centering
\subcaptionbox{LPC}{
\begin{minipage}[t]{0.45\linewidth}
\centering
\includegraphics[width=\linewidth]{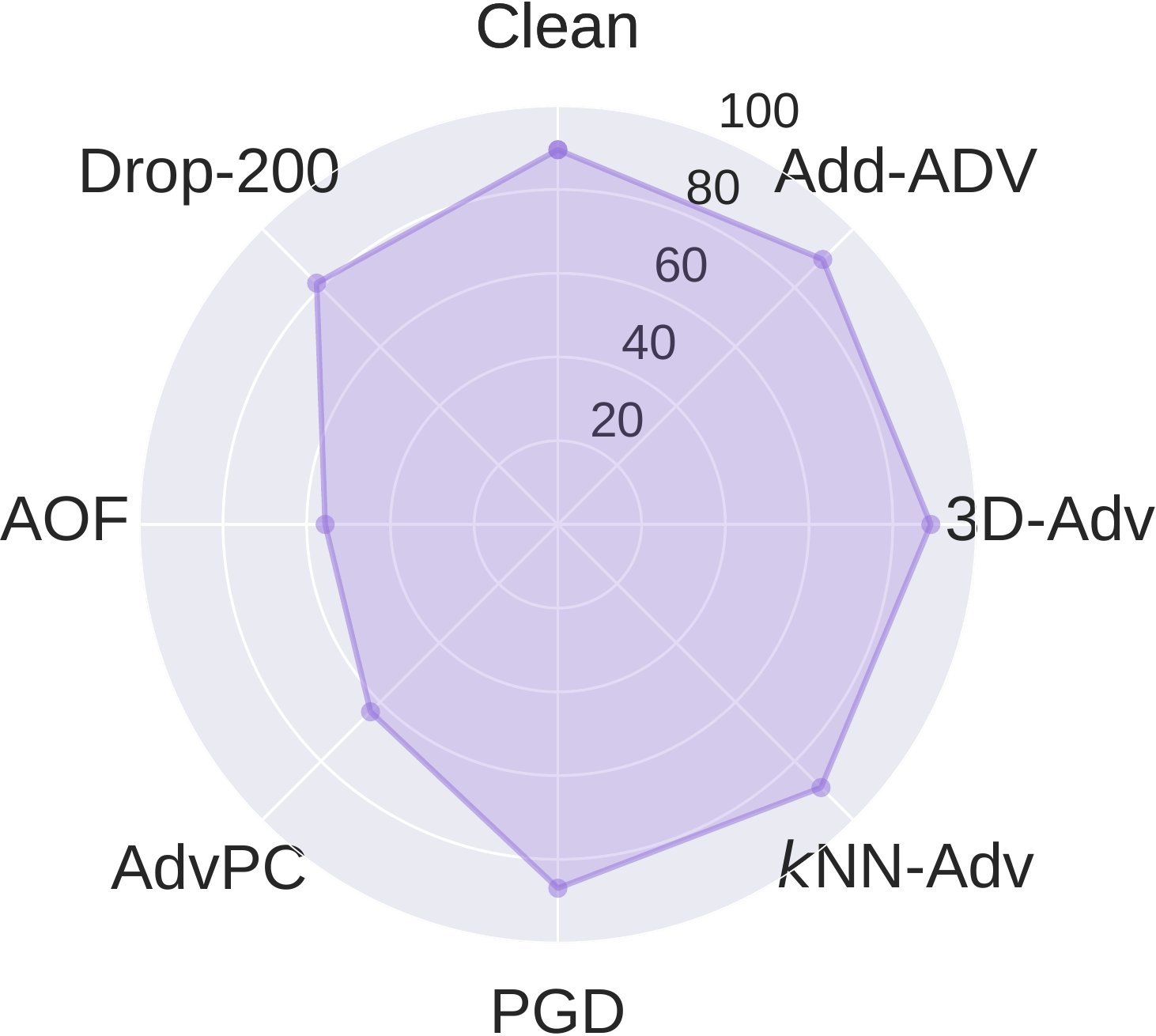}
\end{minipage}%
 }%
\subcaptionbox{Ada3Diff}{
\begin{minipage}[t]{0.45\linewidth}
\centering
\includegraphics[width=\linewidth]{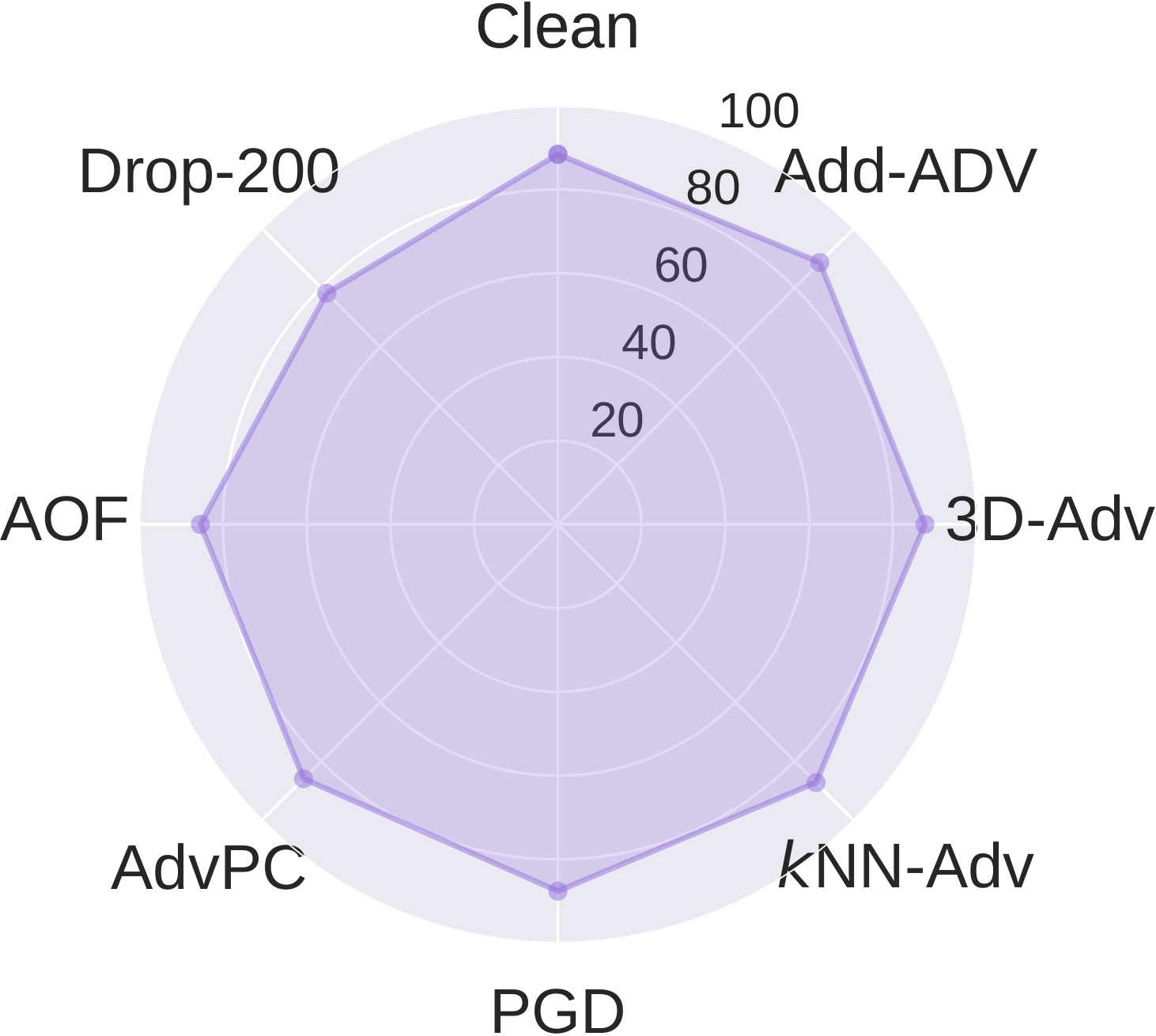}
\end{minipage}%
}%
\centering
\caption{Comparison of the robustness of Ada3Diff and LPC.}
\label{fig:radar}
\end{figure}

\paragraph{More comparison.} 
Table \ref{tab:06} provides the defense results on more models. The proposed approach outperforms present defenses against large distortion attacks while maintaining comparable performance in the face of slight distortions. We also compare Ada3Diff with recent state-of-the-art defense LPC~\cite{LPC}. 
As shown in Figure \ref{fig:radar}, where LPC does not perform well under transfer attacks such as AdvPC and AOF, while the proposed approach has more solid robustness.
\paragraph{Time consumption.}
The average running time of Ada3Diff is calculated on 1 Nvidia RTX2080Ti with a batch size of 1 for a total of 100 samples at 5 and 20 fixed steps. The average time for fixed 5 and 20 timesteps is 0.68s and 2.8s, respectively, and for distortion estimation is 0.2s.

\subsection{Ablation study}
\paragraph{The generalizability of adaptive timestep.}
To evaluate the generalizability of the adaptive step selection method, we compared the differences in timesteps obtained using datasets of different sizes. We computed the mean and standard deviation of the coefficient of variation (COV) between the number of samples and those computed using the entire dataset. Formally, denote $\mathrm S = \{s^i\}, i \in [1,k]$ as the set of the number of samples in each interval, where k is the number of intervals. $\mathrm S_{all}$ is the result over the entire dataset and $\mathrm S_{p}$ represents the set obtained using p$\%$ of the whole dataset. We first calculate the mean and variance of the difference between the two sets as $\mu_i=\frac{1}{2} (s_p^i + s_{all}^i)$, $\sigma_i = \mathbb{E}[(s_p^i - \mu_i)^2 + (s_{all}^i- \mu_i)^2]$. Then the COV in the i-th interval can be computed as COV$_i =  \sigma_i / \mu_i$. Ultimately we calculate the mean and variance of COV for k intervals to reflect the differences between the thresholds obtained for different amounts of data.
Table \ref{tab:cov} demonstrates that similar performance can be obtained with a subset of the dataset as with the complete dataset. 

\paragraph{Analysis of the number of timestep intervals.}
We divided the timestep into more intervals, i.e., 10, and calculated the mean and variance of the COV of the number of samples in each interval for datasets of different sizes. The results in Table \ref{tab:interval} show that dividing more intervals makes the threshold estimation of distortion more sensitive to the size of the dataset.
If the timestep that performs best under each distortion is known, a more detailed interval division strategy would be more helpful, which also requires improvement of the distortion estimation method.

\begin{table}[h!tb]
\setlength\tabcolsep{2pt}
\centering
\caption{Mean/std value of COV ($\times 10^{-2}$). 'uniform-20' represents the dataset sampling 20 samples from each class.}
\label{tab:cov}
\setlength\tabcolsep{2mm}{
\begin{tabular}{l|ccccc}
\toprule[1pt]
        Attacks       & 50\%  & 25\%  & 10\% & uniform-20   \\ \midrule[1pt]
3D-Adv & 0.96/0.65 & 0.85/0.66 & 6.93/2.12  & 4.78/2.82   \\
PGD    & 0.99/9.65 & 0.55/0.61 & 7.39/5.49   & 4.24/1.87   \\
AdvPC  & 1.10/0.89 & 1.35/1.54 & 7.49/6.74 & 6.07/5.85   \\
AOF    & 1.03/1.44 & 1.68/1.68 & 17.4/16.7    & 12.1/14.5 \\
\bottomrule[1pt]
\end{tabular}}
\end{table}

\begin{table}[htbp]
\setlength\tabcolsep{2mm}
\centering
\caption{The mean/std value of COV ($\times 10^{-2}$) for the number of samples in each interval when using 10 intervals. }
\label{tab:interval}
\setlength\tabcolsep{2mm}{
\begin{tabular}{c|ccccc}
\toprule[1pt]
        Scale       & 3D-Adv  & PGD  & AdvPC & AOF   \\ \midrule[1pt]
50$\%$ & 2.07/2.05 & 3.34/2.17 & 3.67/4.34  & 4.01/3.09   \\
25$\%$ & 4.42/3.14 & 3.82/2.11 & 7.70/6.91  & 4.39/3.49   \\
10$\%$    & 9.57/4.26 & 13.6/8.11 & 14.7/21.1   & 19.4/19.6   \\

\bottomrule[1pt]
\end{tabular}}
\end{table}

\subsection{Discussion}
Although point-dropping attacks can be defended to some extent, further improvement is still needed. The dropped point cloud loses some local semantics and is difficult to align with the noisy version of the clean point cloud in the limited diffusion time. We need to use a larger diffusion time to align with the noise distribution of the clean point cloud and to make the reverse denoising preserve local details as much as possible. A promising approach is to use additional conditional guidance, which will be our future work. In addition, instance-level reference-free distortion estimation and more accurate diffusion path selection can further improve robust classification accuracy, which is also the direction of our future exploration. We claim that the adaptive strategy is a necessary component of the defense pipeline and hope it can shed some light on the defense methods based on the diffusion probabilistic model.

\end{document}